\documentclass[a4paper,final,times,fleqn]{cas-dc}

\usepackage[authoryear]{natbib}

\usepackage{amssymb}
\usepackage{amsmath}
\usepackage{latexsym}
\usepackage{booktabs}
\usepackage{url}
\usepackage{hyperref}
\usepackage{caption}
\usepackage{subcaption}
\usepackage{float}

\usepackage{enumitem}
\setlist[itemize]{noitemsep, nolistsep}
\usepackage[export]{adjustbox}

\usepackage{xcolor,colortbl}
\definecolor{Grayl2}{gray}{0.9}
\definecolor{Grayl3}{gray}{0.8}
\definecolor{Grayl4}{gray}{0.7}
\definecolor{Grayl5}{gray}{0.6}
\definecolor{Gray}{gray}{0.5}
\usepackage{array}
\newcolumntype{T}[1]{>{\columncolor{Grayl5}\centering\arraybackslash}p{#1}}
\newcolumntype{Y}[1]{>{\columncolor{Grayl4}\centering\arraybackslash}p{#1}}
\newcolumntype{U}[1]{>{\columncolor{Grayl3}\centering\arraybackslash}p{#1}}
\newcolumntype{I}[1]{>{\columncolor{Grayl2}\centering\arraybackslash}p{#1}}
\newcolumntype{O}[1]{>{\columncolor{Gray}\centering\arraybackslash}p{#1}}
\newcolumntype{P}[1]{>{\centering\arraybackslash}p{#1}}

\begin{document}
\let\WriteBookmarks\relax
\def\floatpagepagefraction{1}
\def\textpagefraction{.001}
\shorttitle{Learning Shape-from-Shading leveraging a realistic synthetic colon database}    
\shortauthors{Ruano J., G\'omez M., Romero E., Manzanera A.}  

\title [mode = title]{Leveraging a realistic synthetic database to learn Shape-from-Shading for estimating the colon depth in colonoscopy images}
\author[1]{Josu\'e Ruano}
\author[2]{Mart\'in G\'omez}
\author[1]{Eduardo Romero} 
\cormark[1]
\ead{edromero@unal.edu.co}
\cortext[1]{Corresponding author: 
  Tel.: (+57 1) 316 5000 Ext. 15025 15183}  
\author[3]{Antoine Manzanera}

\address[1]{Computer Imaging and Medical Applications Laboratory (CIM@LAB), Universidad Nacional de Colombia, 111321, Bogot\'a, Colombia}
\address[2]{Unidad de Gastroenterolog\'ia, Hospital Universitario Nacional, 111321, Bogot\'a, Colombia}
\address[3]{Unit\'e d'Informatique et d'Ing\'enierie des Syst\'emes (U2IS), ENSTA Paris, Institut Polytechnique de Paris, Palaiseau, 91762, Ile de France, France}

\begin{abstract}
Colonoscopy is the choice procedure to diagnose, screening, and treat the colon and rectum cancer, from early detection of small precancerous lesions (polyps), to confirmation of malign masses. However, the high variability of the organ appearance and the complex shape of both the colon wall and structures of interest make this exploration difficult. Learned visuospatial and perceptual abilities mitigate technical limitations in clinical practice by proper estimation of the intestinal depth. This work introduces a novel methodology to estimate colon depth maps in single frames from monocular colonoscopy videos. The generated depth map is inferred from the shading variation of the colon wall with respect to the light source, as learned from a realistic synthetic database. Briefly, a classic convolutional neural network architecture is trained from scratch to estimate the depth map, improving sharp depth estimations in haustral folds and polyps by a custom loss function that minimizes the estimation error in edges and curvatures. The network was trained by a custom synthetic colonoscopy database herein constructed and released, composed of 248\,400 frames (47 videos), with depth annotations at the level of pixels. This collection comprehends 5 subsets of videos with progressively higher levels of visual complexity. Evaluation of the depth estimation with the synthetic database reached a threshold accuracy of 95.65\%, and a mean-RMSE of $0.451 \: cm$, while a qualitative assessment with a real database showed consistent depth estimations, visually evaluated by the expert gastroenterologist coauthoring this paper. Finally, the method achieved competitive performance with respect to another state-of-the-art method using a public synthetic database and comparable results in a set of images with other five state-of-the-art methods. Additionally, three-dimensional reconstructions demonstrated useful approximations of the gastrointestinal tract geometry. Code for reproducing the reported results and the dataset are available at \url{http://cimalab.unal.edu.co/?lang=en&mod=software}.
\end{abstract}

\begin{graphicalabstract}
\includegraphics[scale=0.5]{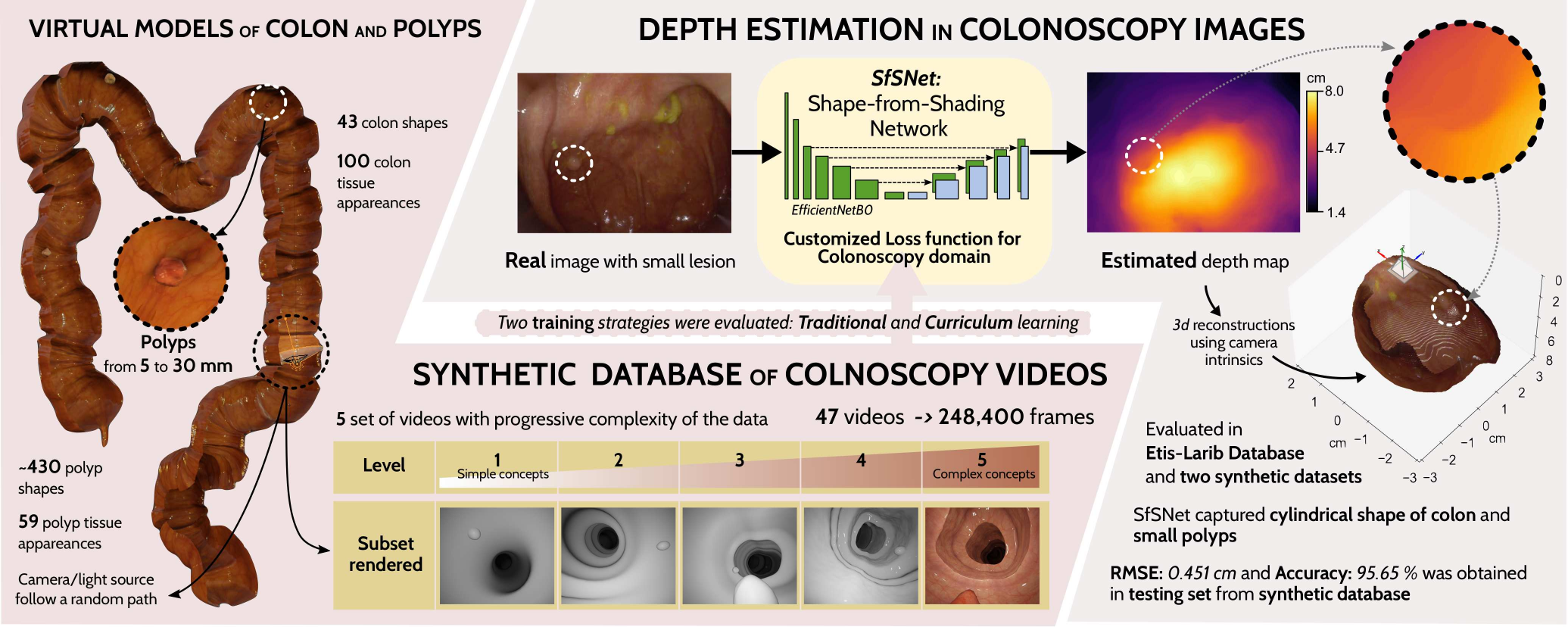}
\end{graphicalabstract}

\begin{highlights}
\item  A convolutional neural network, the SfSNet, estimates a pixel-wise depth map of the colon in a raw frame from the colonoscopy video.

\item A realistic synthetic colonoscopy database with depth annotations that is publicly available, with progressively higher levels of visual complexity.

\item A custom loss function which drives the network by minimizing the estimation error at folds and polyps.

\item Two training strategies are evaluated to optimize the depth estimation in colon folds and precancerous polyps.

\end{highlights}


\maketitle

\section{Introduction}
\label{sec:introduction}

Cancer along the gastrointestinal (GI) tract represents a 23\% of cancer mortality in 2020. Particularly, colorectal cancer (CRC) ranks in 2020 as the second most mortal cancer and the third with the highest incidence worldwide, with an estimate of 1.9 million new cases \citep{globocan2020}. The CRC is usually manifested as polypoid or non-polypoid masses growing from the intestinal mucosa, called neoplastic lesions, which will eventually evolve to adenocarcinoma (malignant) or hyperplasic (benignant) \citep{inoue2003paris}. Early CRC diagnosis is highly challenging because symptoms usually go unperceived, are often non-specific, and most patients remain asymptomatic until advanced stages \citep{sawicki2021,simon2016}. Screening guidelines in average-risk populations contribute to mitigate these incidence and mortality trends \citep{qaseem2019}. Advances in endoscopy and computed tomography have improved detection of precancerous lesions, and for most GI cancers the video colonoscopy is the screening and diagnosis procedure \citep{ladabaum2020}. In a routine colonoscopy, the entire colon and rectum are thoroughly explored for six or seven minutes during the withdrawal \citep{bhurwal2021}. Experts analyze the morphology/shape of the colon wall searching mucosa protrusions as first signs of a lesion, i.e. small depth variations or discontinuities may be perceived as polyps \citep{inoue2003paris,rees2016,amouzeshi2021}. Interestingly, the morphology/shape plays a major role in detecting small ($< 10$ mm) or flat lesions which in most cases look quite similar to mucus \citep{tadepalli2011}. Additionally, for all type of lesions the specialist must remove any observed abnormality, yet the particular procedure depends on the size of the lesion \citep{Izzy2015}: e.g. precancerous or small lesions may be immediately and completely removed (polypectomy) using forceps or wire loops \citep{zhang2018}, whereas huge polyps or large number of polyps require a colectomy or proctocolectomy \citep{lee2009,shussman2014}. Likewise, during the same procedure, a biopsy sample is obtained for later histological analysis to corroborate/compare with the colonoscopy diagnosis and to guide the surveillance of the patient \citep{shussman2014,zhang2018}. 

Colonoscopy is highly operator dependent, i.e. a gastroenterologist requires between 140 to 300 training procedures to gain full competency at detecting lesions \citep{rees2016}. In addition, a colonoscopy may be influenced by adherence to the bowel preparation regime or unrelated physiological conditions (organ motility and liquids). These factors may cause neoplasias to be missed: some studies have reported that 22 to 28 \% of lesions are missed \citep{leufkens2012,Izzy2015}, and the miss rate increases significantly in polyps smaller than 10 mm, which may cause the so-called interval CRC \citep{van2006,kim2017}. Provided that small precancerous lesions are hard to detect during colonoscopy withdrawal even for experienced clinicians, 3d geometry of the colon wall should be useful to highlight small lesions. Indeed, the appearance in terms of color and texture of such small lesions is extremely similar to the surrounding healthy tissue, i.e. fuzzy and open boundaries \citep{antonelli2021,goyal2020,ngoc2020}. In fact the problem results much more difficult due to different noise sources like: a) specular reflections generated by the incident light source  \citep{shandro2020}, b) blurred images produced by frequent shifts or rapid movements during manual operation \citep{wimmer2020,xu2021}, c) occlusion effects by the colonic mucosa, liquids and stools \citep{Taha2017}. In spite of this, several strategies have been proposed to characterize and analyze the morphology of the colon wall during colonoscopy from the estimation of depth. Most relevant works in the state of the art are hereafter presented. 

\subsection{Related works}
\label{subsec:related_works}

Depth estimation in different scenarios, outdoor or indoor, from single or several images, is a major concern for the computer vision community. This problem is ill-posed, but it is also complicated by a set of factors, among others the variable configurations of light sources, the visibility range of the scene, the layout of the objects, and a variety of noise sources \citep{chawla2021}. In case of endoscopy images, interest in estimating the depth of the gastrointestinal tract has been growing in recent years, to visualize and analyze the three-dimensional structure of the tract, estimate the area explored by the specialist, or quantify the volume of structures. Currently, most methods have proposed end-to-end neural networks to recover depth maps from single or pairwise images \citep{hong2014,nadeem2016,mahmood2018,rau2019,freedman2020,mathew2020,itoh2021,cheng2021}.

Several works have estimated depth from a single image, \citet{mahmood2018} approximated the absolute depth frame-by-frame of colonoscopy videos using a two-branch CNN architecture. A superpixel representation fed the first branch responsible for regressing the depth of each region, while the second one smoothed the computed depth using neighboring superpixels. This strategy was trained with synthetic ground-truth depth maps, produced by a virtual colonoscopy (VC) inside digital GI models, i.e. the net was pre-trained with synthetic images from a manually constructed GI tract. For testing, the input image was transformed to a synthetic-like representation by a combination of transformer and discriminator networks, and the estimated depth map generated a three-dimensional reconstruction. In contrast, \citet{rau2019} presented a conditional generative adversarial network (GAN) to estimate the depth of a single image from CT synthetic and real data. The generator or depth estimator is an encoder-decoder CNN, and the discriminator is an encoder that learns to distinguish between fake or real (estimated) depth. This network was trained with a texture-less VC dataset i.e. no color or texture, generated from real CT colonographies, and nine configurations of virtual cameras and light sources.  Another work, by \citet{mathew2020}, conserved the idea of adapting synthetic to real image domain by applying a cycle GAN (composition of two GANs) to generate a texture-less image from a real one and then obtain a depth map. Likewise, \citet{itoh2021} takes advantage of cycle GANs to translate between virtual and real images, adding a Lambertian-reflection model.

On the other hand, several methods have estimated the depth map from multiple images. For example, \citet{freedman2020} estimated the depth and camera pose to find out hidden regions during colonoscopy procedures. First, depth and pose were learned by an unsupervised approach using two CNNs in parallel. The pose CNN computes disparities between contiguous frames, i.e. perception of apparent motion, guiding the second CNN to recover the depth. Then, a per-frame point cloud (three-dimensional GI reconstruction), based on previous estimations, defined a score proportional to the occluded points. Two synthetic databases with depth annotations and one with real images were used to train and test the depth estimation. One of the constructed synthetic collections used pre-fabricated digital models, while the other is a CT-based collection published by \citet{rau2019}. Also, subsets of the previously mentioned collections were annotated by specialists with training and testing scores. A recent method presented by \citet{cheng2021} obtained a per-frame depth map by combining supervised learning in a CNN, and self-supervised learning to ensure a temporal depth map consistency of contiguous frames by estimating the optical flow.

Although state-of-the-art results look promising, depth estimation is still an open problem. Additionally, the lack of publicly available and highly-variable datasets is a limitation for training, testing, and benchmarking with existing methods. 

\subsection{Contributions}

A mental depth estimation, a three-dimensional model of the GI tract, is constructed by specialists during a colonoscopy as a guide of the whole procedure \cite{Soetikno2021}. Specifically, precancerous or small lesions are usually observable when subtle maneuvers of the device triggers mental differences of the depth perception. The main contribution of the present work consists in estimating colon depth from gradual luminance variation produced by the light source of the colonoscopy device. Specific contributions are:

\begin{itemize}
    \item  A convolutional neural network, the SfSNet, specialized in estimating a pixel-wise depth map from the raw colonoscopy image. A custom loss function which drives the network by minimizing the estimation error at edges and curvatures, i.e., haustral folds and polyps.
    \item A novel synthetic colonoscopy database with depth annotations is released. This database is composed of five sets of videos with progressively higher levels of complexity in terms of appearance and morphology.
    \item Evaluation of two training strategies applied to train the SfSNet, namely traditional (TL) and curriculum learning (CL).
\end{itemize}

\section{Methodology}\label{sec:methodology_sec}

\begin{figure*}[ht!]
\centering\medskip
\includegraphics[scale=0.95]{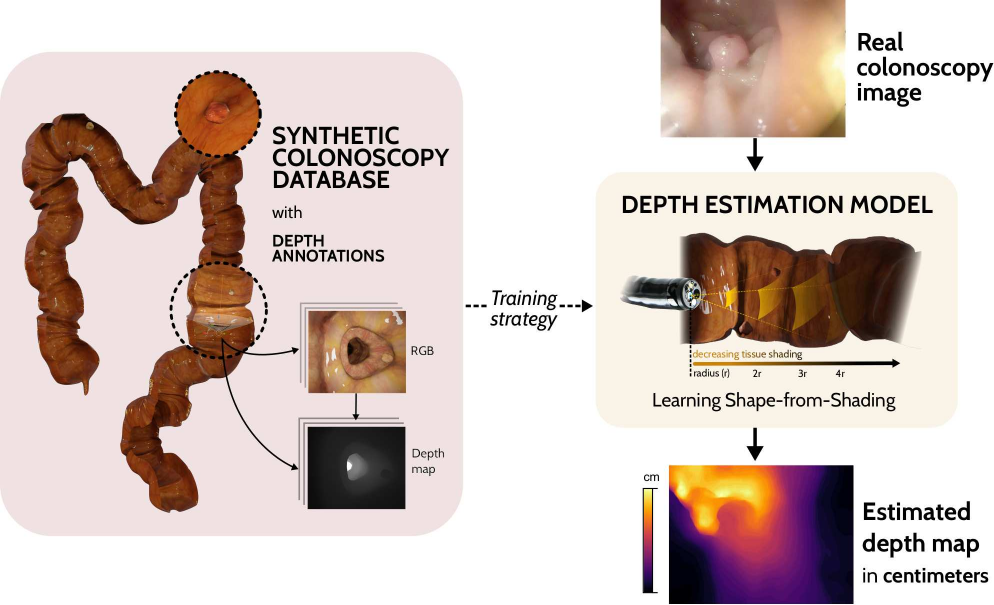}
\caption{Pipeline of the proposed approach. First, a synthetic colonoscopy database is constructed (see Section \ref{subsec:synthetic_colon_db}). Then, a Shape-from-Shading strategy estimates depth map (see Section \ref{subsec:depth_estimation}). Finally, a depth estimation model trained in synthetic data is applied in real colonoscopy images.}
\label{fig:pipeline}
\vspace{-0.3cm}
\end{figure*}

The colonoscopy device produces a variable illumination of the different colonic structures as a function of both the distance and orientation to the camera-light source. The depth of each pixel, i.e., the radial distance between the camera center and a 3d position of the colon wall, in the obtained image may be associated with intensity values, assuming the surface is Lambertian, and in a lesser proportion, specular. Hence, this work presents a strategy (see Section \ref{subsec:depth_estimation}) to estimate a depth map from a single colonoscopy image by learning the variations of shading with respect to both the distance and orientation of the camera, a technique known as shape-from-shading. Herein shape corresponds to the $3d$ structure of the colon captured by the depth map  \citep{zhang1999shape,ramachandran1988perception, ikeuchi1981numerical}. In a colonic exploration, these shapes are function of a light source which dims out as the distance increases by an intensity attenuation model, an intensity inverse square law, $I \propto \frac{1}{r^{2}}$, where $I$ is the intensity, and $r$ represents the radial depth from the light source \citep{prados2005}.

Classic convolutional neural networks are used to learn these variations under a supervised approach, specifically DenseNet169 \citep{huang2017}, ResNet50 \citep{he2016deep} and EfficientNetB0 \citep{tan2019}. In addition, a custom loss function minimizes the depth estimation errors in curved shapes, i.e., haustral folds and polyps. These networks are trained using realistic synthetic videos of colonoscopy, since real colonoscopy videos with depth annotations are not available (see Section~\ref{subsec:synthetic_colon_db}). A synthetic video is generated from a virtual environment, which emulates three-dimensional colon structures and polyps by mapping extracted visual patterns from real images to three-dimensional ($3d$) meshes constructed from averaged colon measurements, specifically from a study with 505 computed tomography colonographies of adult subjects \citep{khashab2009}. The virtual environment simulates an actual colonoscopy withdrawal, which lasts 6 minutes in average, following a random path with 3 degrees of freedom. During this virtual exploration, RGB synthetic frames with dense ground truth depth maps are generated with the camera as the reference point.

\subsection{Colonoscopy depth estimation by Shape-from-Shading}
\label{subsec:depth_estimation}

A depth map from a single colonoscopy frame refers to the estimated radial distance between the camera center and the colon wall. Depth estimation, based on shading, is an ill-posed problem simply because the 2d-image is a projection of the $3d$-scene and multiple approximations co-exist \citep{zhang1999shape,ramachandran1988perception, ikeuchi1981numerical, bhoi2019monocular}. Different constraints regularize the solution of the problem, namely (a) the camera is modeled as a pinhole with constant focal length, (b) position of the light source is practically the same as the optical center of the camera, and (c) reflectance of the mucosa is assumed to be Lambertian in a large proportion \citep{horn1986, prados2005,itoh2021}, which implies that albedo is mostly uniform \citep{itoh2021}. These constraints comply with reflected light decreasing according to
$L \propto R \frac{\mathbf{w}.\mathbf{n}}{r^2}$ where $L$ is the reflected intensity, $R$ is the surface reflectance, $\mathbf{w}$ is the light vector, $\mathbf{n}$ is the normal of the surface and $r$ is the radial distance \citep{prados2005,okatani1997,itoh2021}. Hence, a machine learning framework may learn this simple relation/model, i.e., depth estimation in colonoscopy may be reliably obtained given a single RGB image by a pixel-level regression task. Previous works have shown coherent depth estimations obtained by a regression-based strategy with similar restrictions to those ones herein used  \citep{itoh2021,mathew2020,mahmood2018,rau2019}. In this work, such regression is performed by supervised learning of a fully convolutional network-based (FCN) strategy, herein named Shape-from-Shading Network (SfSNet). The details of the network are presented below, and the overview is shown in Figure~\ref{fig:cnn}.

\subsubsection{Architecture of the convolutional network}
\label{subsubsec:network_architecture}

\begin{figure*}[ht!]
 \centering\medskip
 \includegraphics[scale=0.85]{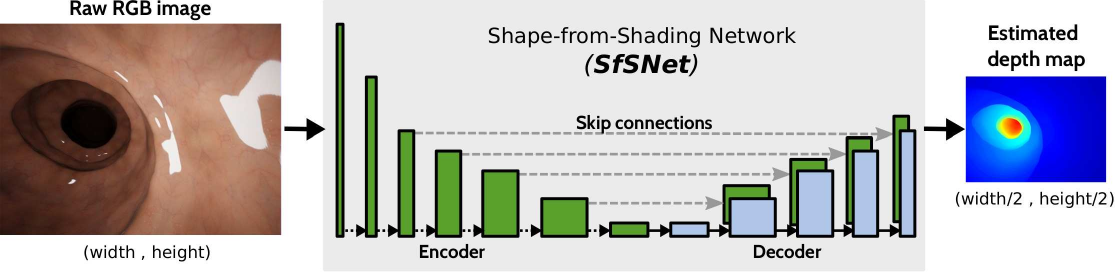}
 \caption{Overview of the convolutional network, built as an encoder-decoder architecture with skip connections. An RGB input image is used to apply a pixel-wise regression to obtain a scalar depth map with half the spatial resolution.}
 \label{fig:cnn}
 \vspace{-0.3cm}
\end{figure*}

Currently complex CNN architectures estimate depth in natural and indoor scenarios, i.e., multiple light sources and non-Lambertian surfaces. In colonoscopy, a unique light source and Lambertian constraints model the scene, reason why a simple CNN configuration might approximate depth maps, i.e., a standard encoder-decoder architecture with long skip connections \citep{alhashim2018high}. This CNN has been widely validated for segmentation and regression tasks, in particular problems in which regions of interest are shapes with borders since long skip connections have been documented to preserve high-frequency details (e.g. borders of haustral folds and polyps) \citep{drozdzal2016,dosovitskiy2015}. The long skip connections refer to use encoder feature maps in the decoding phase (upsampling), integrating detail information lost during downsampling \citep{drozdzal2016,dosovitskiy2015}. Specifically:

\begin{itemize}
    \item \textbf{Encoder:} Three conventional but yet representative state-of-the-art feature extractors are explored. The architectures herein used are DenseNet169 \citep{huang2017}, ResNet50 \citep{he2016deep} and EfficientNetB0 \citep{tan2019}, with different number of layers and parameters.
    
    \item \textbf{Decoder:} In regression tasks, the up-sampling network or decoder is constructed as the reverse architecture of the encoder. However, some works in the state of the art showed that a simpler version of decoder achieves a performance very similar compared to a decoder with equal complexity as the encoder, leading to asymmetrical architectures \citep{alhashim2018high,kim2019asymmetric,kazerouni2021ghost}. Additionally, those works demonstrated that up-sampling may output an image with smaller spatial resolution, being half the size enough for a high-detailed image \citep{shi2021j,alhashim2018high}. For these reasons, the decoder is formed only by four up-sampling stages. Each stage is composed of a bi-linear up-sampling layer along with its respective skip connection from the encoder and followed by two convolutional layers. Finally, the encoder yields a scalar depth image with half of the RGB size input image.
\end{itemize}

\subsubsection{Customized loss function}
\label{subsubsec:custom_loss}
Usually protruding structures in colon topography, polyps, and haustral folds, are highly similar in terms of color and texture when they are observed in two dimensions. These structures however can be more easily distinguishable by including local $3d$ geometry captured in the depth map. Polyps and haustral folds, basically curved shapes, show characteristic gradients in the depth map (first-order derivatives) and especially in the depth curvatures (second-order derivatives)\citep{Hanbay2018,Zhao2006}.

For those reasons, the loss function $\mathbf{L}_{SfSNet}$ herein proposed is composed of three weighted terms to balance between a smooth reconstruction of the depth map with the $L_z$ term, while also preserving the high-frequency details or borders with $L_e$ and $L_c$, respectively related to the edge (gradient) and curvature of the depth maps. The loss function finally evaluates the estimated depth $\hat{d}$ with respect to the ground truth $d$ and is defined as:

\begin{equation}
\mathbf{L}_{SfSNet}(d,\hat{d}) = w_{1} L_z(d,\hat{d}) + w_{2} L_e(d,\hat{d}) + w_{3} L_c(d,\hat{d})
\label{eq:global_loss}
\end{equation}

\begin{itemize}
    \item $L_z$ term evaluates the depth estimation in the complete colon wall. This is the pixel-wise $\mathcal{L}_1$ loss defined over the depth image: 
     \begin{equation}
    L_z(d,\hat{d}) = \frac{1}{n} \sum \limits_{p} |d_{p} - \hat{d_{p}}|
\label{eq:loss0}
    \end{equation}
    where $p$ denotes each pixel and $n$ the number of pixels.
    \item $L_e$ term minimizes the depth estimation errors of structures with borders. It is the $\mathcal{L}_1$ loss defined for the depth image gradients:
    \begin{equation}
    L_e(d,\hat{d}) = \frac{1}{n} \sum\limits_{p} |\mathbf{g}_{x}(d_{p}) - \mathbf{g}_{x}(\hat{d_{p}})| + |\mathbf{g}_{y}(d_{p}) - \mathbf{g}_{y}(\hat{d_{p}})|
    \label{eq:loss1}
     \end{equation}
     where $\mathbf{g}_{x}$ and $\mathbf{g}_{y}$ respectively denote the horizontal and vertical components of the gradient.
    \item $L_c$ term is proposed to minimize the depth estimation errors of curved borders from polyps and haustral folds. It is the $\mathcal{L}_1$ loss defined for second-order partial derivatives $H$ of depth image, i.e., Hessian derivatives, which are well-known in different medical image domains, including colonoscopy, to capture curvature
    \citep{Iwahori2015,Hanbay2018,Zhao2006}.
    \begin{equation}
    \begin{aligned}
    L_c(d,\hat{d}) = \frac{1}{n} \sum_{p} |\mathbf{H}_{xx}(d_{p},\sigma) - \mathbf{H}_{xx}(\hat{d_{p}},\sigma)| + \\ 2 |\mathbf{H}_{xy}(d_{p},\sigma) - \mathbf{H}_{xy}(\hat{d_{p}},\sigma)| + |\mathbf{H}_{yy}(d_{p},\sigma) - \mathbf{H}_{yy}(\hat{d_{p},\sigma})|
    \end{aligned}
    \label{eq:loss2}
    \end{equation}
    where $\mathbf{H}_{xx}$, $\mathbf{H}_{xy}$ and $\mathbf{H}_{yy}$ denote the second-order partial derivatives with a standard deviation $\sigma$.
    \item $\sigma$ impacts the size of Hessian kernels which was manually tunned by visually evaluating that soft borders of small lesions and strong borders of haustral folds were captured. The value found is $\sigma=3$ corresponding to a kernel size of $12 \times 12$.
    \item $w_{1}$, $w_{2}$ and $w_{3}$ are the weight parameters to balance the loss function $\mathbf{L}_{SfSNet}$.  The values are tuned by a grid search, in which the weights were chosen to penalize to a greater extent the depth estimation in polyps and haustral folds by greater values in $w_{2}$ and $w_{3}$. The values used in the experiments were $w_{1}=0.1$, $w_{2}=0.3$ and $w_{3}=0.6$. 
\end{itemize}

It is worth mentioning that $L_c$ and $L_e$ terms are calculated on depth maps and not on color images, to also help minimize the depth estimation error in flat photometrically structured areas like vascularity, specular reflections, or malignant tissue textures.  In addition, an ablation study is performed to demonstrate the impact of each term in the loss function and the selected weight parameters, further details are given in Section~\ref{subsubsec:loss_ablation_study}.


\subsubsection{Learning and inference strategies}
\label{subsubsec:learning_strategies}

Optimization algorithms of neural networks may get stuck in a local minimum during the process of updating the weight parameters. Being deep learning a pure data-driven strategy, an optimum is reached by exhaustive training with a large quantity of data and as much variability as possible. Nevertheless, availability of annotated open-access medical data is limited, much more in case of colonoscopy with depth estimations. Although depth annotated synthetic data is an option to improve models' performance, synthetic data would hardly provide the variability of real data in terms of color, texture, anatomic shape, and noise sources. A reliable alternative consists in guiding the learning process by progressively including data with higher and higher levels of complexity, even though data variability is still limited, in other words, setting variability along the different training stages. This work then evaluates two training workflows for the proposed network: first, a traditional strategy, and second, the proposed method that takes advantage of transfer learning \citep{tan2018survey} in several stages, i.e.,  Curriculum learning. These approaches are detailed hereafter.\medskip
\begin{itemize}
    \item \textbf{\textit{Traditional learning strategy:}}    
    This network is trained as state-of-the-art deep learning frameworks usually do, i.e., in a single training stage and several epochs. Although these frameworks generally apply transfer learning, here the network is trained from scratch since no pre-trained end-to-end networks are available to estimate depth. Specifically, network weights are randomly initialized and typical regularization strategies (for further details see Experimental setup section) are used to avoid overfitting. This process will be referred hereafter as Traditional Learning (TL) strategy and an overview is shown in Figure~\ref{fig:traditional_learning}.
    \begin{figure}[h]
        \centering
        \includegraphics[scale=0.9]{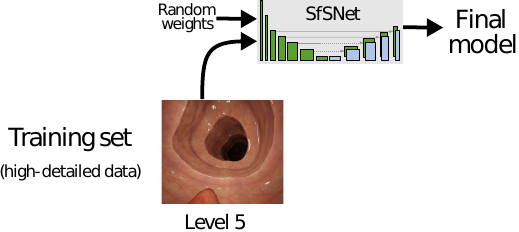}
        \caption{Traditional learning strategy}
        \label{fig:traditional_learning}
    \end{figure}
    \item \textbf{\textit{Curriculum learning strategy:}}
    \begin{figure*}[t]
        \centering
        \includegraphics[scale=0.9]{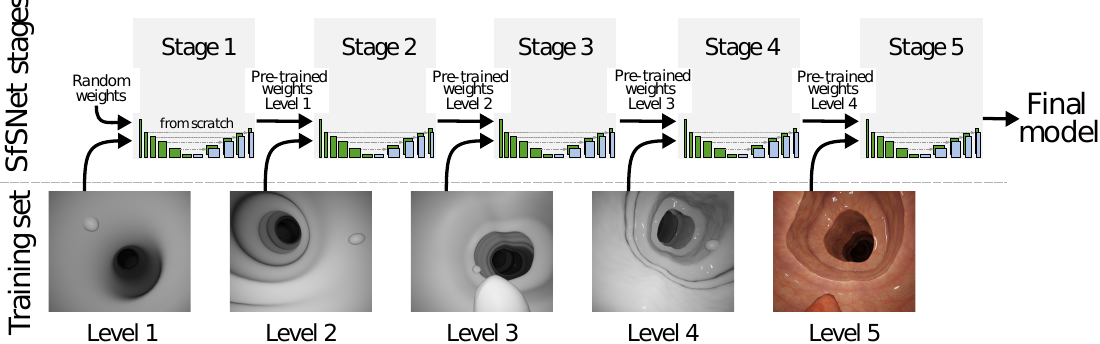}
        \caption{Overview of curriculum learning strategy performed at five stages, with five subsets of synthetic colonoscopy data.}
        \label{fig:curriculum}
    \end{figure*}
    The basic idea of a Curriculum Learning (CL) is to ensure that the found solution follows the global extremum by gradually complexifying the parameter space while improving generalization and speeding up convergence  \citep{parisi2019,bengio2009curriculum,narvekar2020,graves2017automated,hacohen2019}. Essentially, this process involves training a network in several stages with samples in a meaningful order, i.e., the learning process starts from data expressing the global relationships and progressively increasing the detail  \citep{wu2017online,Surendranath2018}. In colonoscopy, CL reinforced learning of specific concepts, e.g., the SfSNet was trained by a staged learning strategy focused on accurate estimation of polyps and haustral folds, starting from images with simplified structures, followed by typical color and textural patterns of real colon appearance. Nevertheless, as no public colonoscopy database is available with depth annotations, even less with progressive levels of detail, a synthetic colonoscopy database with depth annotations was herein constructed. Further details of the synthetic database are presented in the section \ref{subsec:synthetic_colon_db}.
    
    As previously said, the SfSNet was progressively trained in five stages, meaning the trained network in one stage is used in the next stage, as shown in Figure~\ref{fig:curriculum}, starting from scratch and with randomly initialized weights. In the first stage, the SfSNet is trained with a dataset (Level 1 of the synthetic database) obtained by modeling the colon as a bent cylindrical structure without haustral folds, and polyps as spherical structures. 
    
    Once training and validation losses stabilize, typically around zero, the obtained weights of the network feed a second and more complex stage (Level 2 of the synthetic database), i.e. haustral folds are incorporated to the colon and rectum model. This change of training and validation data is repeated five times, adding details as follows: in stage three (Level 3 of the synthetic database), the lumen shape's model assumes the morphology of the different colon segments and polyps take deformed spherical shapes; stage four (Level 4 of the synthetic database) integrates typical colon and polyp textures; and in stage five (Level 5 of the synthetic database)  variability of real colonoscopies are mapped to the model and integrated, i.e., colon and polyp models include shapes, colors, and textures as close as possible to real structures. The effect of this training protocol is that the problem is approximated by a set of models, each dealing with progressively higher levels of complexity. 
    
    Actually, the CL strategy amounts to apply a sequence of homogeneous transfer learning stages. The set of obtained transformations  $\mathcal{D}$ is defined as $\mathcal{D} = \{ \mathbf{F}_{k}^{s},P(\mathbf{F}_{k}^{s}) \}$, being $\mathbf{F}_{k}^{s}$ the feature space in a CL stage $s$, $k$ the used architecture (DenseNet169, ResNet50 or EfficientNetB0), and $P$ the marginal probability distribution. The feature space $\mathbf{F}$ is defined by the depth estimation task, $\mathcal{T}_{d} = \{\mathbf{Y}, P(\mathbf{Y}|\mathbf{F}_{k}^{s})\}$, i.e., the task $\mathcal{T}_{d}^{s}$ determines the depth regression weights $Y_{d}$ for every CL stage and the transfer learning codes the next stage $\mathcal{T}_{d}^{s+1}$, as $\mathcal{T}_{d}^{s} = \{ Y_{d}, \mathbf{F}_{k}^{s} \} \to \mathcal{T}_{d}^{s+1} = \{ Y_{d}, \mathbf{F}_{k}^{s+1} \}$ \citep{tan2018survey}. This process is iterated five times, up to the last Level of synthetic data. Summarizing:
    
    \begin{equation}
    \mathcal{T}_{d}^{s1} \rightarrow \mathcal{T}_{d}^{s2} \rightarrow \mathcal{T}_{d}^{s3}\rightarrow \mathcal{T}_{d}^{s4} \rightarrow \mathcal{T}_{d}^{s5}
    \end{equation}
    
    being $s1$, $s2$, $s3$, $s4$, and $s5$ each stage of CL, respectively. Finally, a depth map model is applied in real colonoscopy images.
    
    \end{itemize}

\subsection{Construction of the synthetic colonoscopy database}
\label{subsec:synthetic_colon_db}

The synthetic database was iteratively constructed with \textit{Blender}, an open-source computer graphic software, since it facilitates: (a) build complex 3d shapes such as colon segments or polyps, (b) emulate small ($\approx 5mm $ of diameter) and large environments captured by wide-angle camera and light sources, challenges of a typical colonoscopy with narrow colonic lumen, and (c) set up those elements in a real-world coordinate system. Moreover, \textit{Blender} provides a module to develop custom scripts to adjust the camera/light and $3d$ model parameters with any configuration, e.g., realistic colon exploration requires particular camera/light paths inside the colon or variable size and frequency of haustral folds, among others. This is not a minor issue, some authors for instance \citep{mahmood2018,freedman2020} have used $Blender$ to build simpler synthetic colonoscopy datasets, v.g., \citet{mahmood2018} developed colon and polyp models with no texture or color, and  videos rendered by Freedman \citep{freedman2020} showed no specular reflection. Further database details are hereafter provided. 

\subsubsection{Basic colon and polyp models}
\label{subsubsec:colon_polyp_base}

\begin{figure}[h!]
 \centering
\includegraphics[scale=0.75]{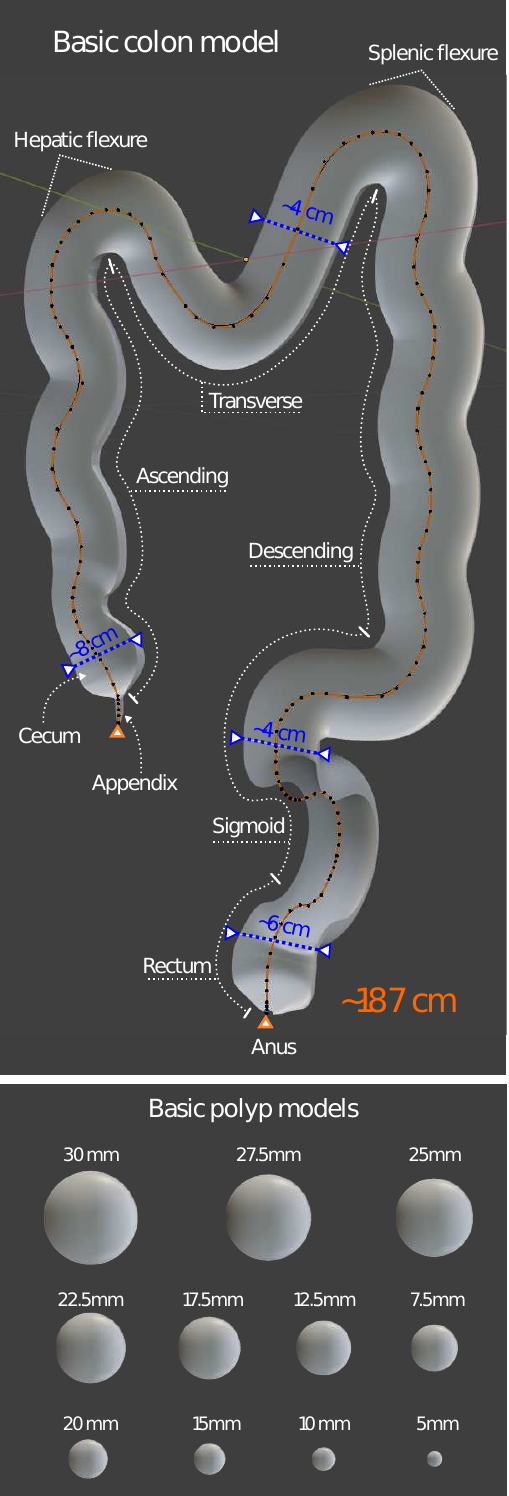}
\caption{A structure matching anatomical colon measures and a set of spheres with sizes from $30 \: mm$ to $5 \: mm$ are the basic colon and polyp models, respectively.}
\label{fig:base_3d_models}
\end{figure}

The colon model baseline was, under a Lambertian frame, a deformed cylindrical structure, while polyps were included as spherical protrusions, with no texture. Overall, the colon model was built using reported measurements of length, diameter, and angles of the low-GI tract from a study with 505 computed tomography colonoscopies of adult subjects \citep{khashab2009}. The process starts in a $3d$ reference frame by manually setting a set of points and interpolating a continuous curve (Bézier curve), matching reported measurements \citep{khashab2009} of the ascending ($~ 30 \: cm$), transverse ($~ 58 \: cm$), and descending ($~ 30 \: cm$) colon segments, as well as the sigmoid ($~ 49 \: cm$) and rectum ($~ 20 \: cm$) (see yellow line and red dots in the top panel of Figure~\ref{fig:base_3d_models}). Angles of right and left flexures were less than $90$ degrees, as reported in the literature \citep{wozniak2022,khashab2009}, specifically the hepatic and splenic flexures were $~ 40^\circ$ and $~ 30^\circ$, respectively. The final curve length reaches about $187 \: cm$ from which a cylindrical mesh was generated. The circular shape of the cylinder is given by a circular curve. The initial cylinder, with the same diameter, is locally deformed to match reported diameters~\citep{khashab2009} of the low-GI tract (see top panel of Figure~\ref{fig:base_3d_models}). Likewise, seven basic polyp spherical models varied their diameters from 5 to 30 mm (see bottom panel of Figure~\ref{fig:base_3d_models}). These basic colon and polyp models are complexified in the following steps.

\subsubsection{Emulating a realistic colon and polyp appearance}
\label{subsubsec:modifications}

A realistic appearance is achieved by transforming the colon and polyp basic features in terms of (a) haustral fold, (b) deformed colon model, (c) surface, (d) colon texture and (e) deformed polyp which is also modified along with (c) and (d). Specific modifications are further explained hereafter.

\begin{itemize}

\item \textbf{\textit{a. Haustral fold}}: haustral folds are visually distinguished as structures protruding the colonic mucosa, corresponding to circumferential contractions of the inner muscular layer of the colon \citep{huizinga2021}, herein simulated by random selection of points from the Bézier curve and separated by 3 to $6 \: cm$  \citep{huizinga2021}. Provided diameter folds vary by the colon motility or air insufflation from the colonoscope, folds ($45 \pm 15$ per model)  randomly vary their diameter between $2.8$ and $7.5 \: cm$. Specifically, a fold corresponds to a set of connected mesh vertices which are shrunk to match a smaller transversal cross-section of the colon. Afterwards, any surface discontinuity generated by this folding process is smoothed out.

\item \textbf{\textit{b. Deformed colon model}}: The transversal colon cross-section or lumen shape is different for each segment, i.e., ascending, descending, and transversal colon segments show triangular, oval, or circular shapes, respectively. The set of $38$ possible lumen shapes was collected from real colonoscopy videos (dataset described in \ref{sec:real_dataset}).

\item \textbf{\textit{c. Inner colon and polyp irregularities}}: 
A real colonic appearance requires an irregular surface, with an asymmetric distribution of the light within the cavity. Overall the organ motion is not homogeneous and certain regions of the colonic wall are hypertrophied as a result of the asymmetric feces motion. Simulation of this condition is achieved by a noise-based pattern, the $3d$ Perlin noise, which closely resembles these irregularities. Additionally, the surface is configured to produce specular reflections (well-defined white regions) in specific positions, where the optical axis of the camera is aligned with the surface normal during the video rendering. On the other hand, a synthetic surface with malignancy degrees imitates the polyp aspect, i.e., the more malign the polyp, the more irregular the polyp surface. This is achieved by a Voronoi texture, which looks like abnormal tissue and makes the effect more evident in larger polyps ($30 \: mm$). The specular reflections were also applied to the polyp models.

\item \textbf{\textit{d. Tissue appearance}}: Depending on the colon segment and malignancy degree, appearance of the colon mucus and polyp tissue varies. A combination of vessels, crypts, liquids, and specular reflections produces complex texture patterns which are difficult to imitate. Therefore, small patches were extracted from colonoscopy videos, specifically from colon structures and polyps, to be integrated in the simulation scenario. This process requires choosing a homogeneous shading frame, i.e. from a region with similar depth. However, as mentioned before, surface irregularities result in undesired specular reflections which must be filtered out by a classic in-painting method to ensure reflections originate from the synthetic surface during the video rendering (as explained in (c) inner colon and polyp irregularities). Afterwards, the patch was resized so that it looks similar to the visually estimated area from the actual colonoscopy. Finally, the virtual model is painted by mirroring and tiling the patch as long as needed to cover the entire virtual surface.

During video rendering, the tissue appearance of colon and polyp models is randomly changed every 3 frames, to increase even more variability for training and testing of the methods proposed. 100 patches of colon tissues and 59 patches of polyps were extracted in 97 videos from different subjects (further details of this database are explained in Section \ref{sec:real_dataset}) to generate the set of tissue appearances.

\item \textbf{\textit{e. Deformed polyp}}: Polypoid forms with a size less than $30 \: mm$ may appear like irregular elliptical or spherical protrusions \citep{inoue2003paris}. This condition is mimicked by modifying the shape of a polyp basic model (sphere), by randomly approaching or moving away the vertex mesh from the center of the sphere (radius), constrained to $\pm 10\%$ of the radius.

\end{itemize}

A colon basic model and a set of polyp basic models ($10 \pm 2$ lesions) can then mimic a large set of real scenarios by combining all these mentioned modifications, providing a synthetic structure with a high level of detail.

\subsubsection{Camera and light source configuration}
\label{subsubsec:camera_light}

In a routine colonoscopy procedure, the gastroenterologist starts when the tip of the colonoscope reaches the ileocecal valve, and then she/he slowly withdraws it while exploring the tract. A virtual camera was configured to simulate the typical colonoscope with a wide-angle lens of 110 degrees of view. Next to the virtual camera, a virtual light source emits a cone-shaped beam with a radius of 140 degrees. These two virtual elements follow a path within the virtual colon model, which avoids them to collide with haustral folds or large polyps. This path is generated by randomly displacing each point of the Bézier curve (element defined in the colon basic model) within a circumference in the orthogonal plane with a $1 \: cm$ radius. The camera and light source align their optical axes with the slope formed between two consecutive points of the Bézier curve.

\subsubsection{Video rendering} 
\label{subsubsec:video}

A synthetic video is rendered (frames per second and spatial resolution in section~\ref{sec:synthetic_dataset}) as long as the camera and light source explore a colon model with a set of polyps. Camera/light path, colon, and polyp models are parameterized as explained in the previous section. Each synthetic RGB frame is generated together with a depth map at the same spatial resolution, i.e. a scalar image in which each pixel represents the absolute distance in $cm$ between the colon point and the camera plane.


\subsubsection{Curriculum learning strategy with virtual videos} 
\label{subsubsec:subsets-curriculum}

Five subsets of videos were generated by level of detail, i.e., level one is the simplest subset, and more details are gradually included until level five. Each level is described in Table~\ref{table:modifications_synthetic}.

\begin{table*}[!t]
\renewcommand{\arraystretch}{1}
\begin{tabular}{O{0.3cm}P{2.1cm}I{2.1cm}U{2.5cm}Y{3.6cm}T{4.2cm}}
  & \textbf{Level 1} & \textbf{Level 2} &  \textbf{Level 3} & \textbf{Level 4} &  \textbf{Level 5} \\ 
  
  {\rotatebox[origin=rc]{90}{\textbf{Colon model}}}
  & \small 
  \begin{itemize}[leftmargin=0.1cm,partopsep=0pt,labelsep=0.1cm] 
  \item Basic model
  \end{itemize} 
  & {\small 
  \begin{itemize}[leftmargin=0.1cm,partopsep=0pt,labelsep=0.1cm] 
  \item Haustral folds
  \end{itemize}} 
  & {\small 
  \begin{itemize}[leftmargin=0.1cm,partopsep=0pt,labelsep=0.1cm] 
  \item Haustral folds
  \item Deformed model
  \end{itemize}}
  & {\small 
  \begin{itemize}[leftmargin=0.1cm,partopsep=0pt,labelsep=0.1cm] 
  \item Haustral folds 
  \item Deformed model  
  \item Inner irregularities
  \item Specular reflections
  \end{itemize}}
  & {\small
  \begin{itemize}[leftmargin=0.1cm,partopsep=0pt,labelsep=0.1cm] 
  \item Haustral folds
  \item Deformed model
  \item Inner irregularities 
  \item Specular reflections
  \item Tissue appearance (texture)
  \end{itemize}}\\
  
  {\rotatebox[origin=rc]{90}{\textbf{Polyp model}}}
  & {\small 
  \begin{itemize}[leftmargin=0.1cm,partopsep=0pt,labelsep=0.1cm]  
  \item Basic model
  \end{itemize}}
  & {\small 
  \begin{itemize}[leftmargin=0.1cm,partopsep=0pt,labelsep=0.1cm]  
  \item Basic model
  \end{itemize}}
  & {\small 
  \begin{itemize}[leftmargin=0.1cm,partopsep=0pt,labelsep=0.1cm]  
  \item Deformed polyp
  \end{itemize}}
  & {\small  
  \begin{itemize}[leftmargin=0.1cm,partopsep=0pt,labelsep=0.1cm] 
  \item Deformed polyp
  \item Surface irregularities
  \item Specular reflections 
  \end{itemize}}
  & {\small 
  \begin{itemize}[leftmargin=0.1cm,partopsep=0pt,labelsep=0.1cm]   
  \item Deformed polyp
  \item Surface irregularities
  \item Specular reflections 
  \item Tissue appearance (texture)
  \end{itemize}}\\
  
\end{tabular}
\caption{Colon and polyp model descriptions of the five subsets from the synthetic colonoscopy database.}
\label{table:modifications_synthetic}
\end{table*}

\subsection{Datasets}
\subsubsection{Synthetic colonoscopy database}
\label{sec:synthetic_dataset}
A total of 47 synthetic videos at a spatial resolution of $1\,280 \times 1\,080$ pixels was rendered at 15 frames per second. Each video is configured to last 6 minutes since it corresponds a complete bowel exploration, i.e., displacing the camera/light from the caecum to the anus of the colon model. A video is typically composed of $5\,400$ RGB frames and hence the complete database contains $248\,400$ frames. For each RGB frame, an associated depth map is a gray-scale image representing depth values between $0 \text{ to } 25 \: cm$, being this distance the farthest one observed during colonoscopy, approximately half of the reported length of the longest segment of the colon, i.e., the transverse. Additionally, a gamma correction is applied to the depth maps for training and testing  models, aiming to enhance the representation of low and intermediate distances ($< 15 \: cm$). In this range the colon wall is better illuminated, and accurate depth estimation may improve detection of abnormalities, unlike darker regions ($>15 \: cm$) where abnormalities are hardly observed. 

\begin{table}[ht]
\small
\centering
\renewcommand{\arraystretch}{1.2}
\begin{tabular}{cccccc}
\toprule
\multicolumn{5}{c}{\textbf{\#
Videos per Level of detail}} & \multirow{2}{*}{\textbf{Total of videos {\footnotesize(frames)}}} \\ \cline{1-5}
\textbf{L1} & \textbf{L2} & \textbf{L3} & \textbf{L4} & \textbf{L5} &
\\ \hline
5 & 5 & 5 & 5 & 27 & \begin{tabular}[c]{@{}c@{}} 47 {\footnotesize(248\,400)}\\ {\footnotesize(5\,400 per video)}\end{tabular} 
\\ \toprule
\end{tabular}
\caption{Description of the synthetic database in subsets.
}
\label{table:synthetic_database.}
\end{table}

\subsubsection{Real colonoscopy databases \label{sec:real_dataset}}
\begin{itemize}

    \item \textbf{\textit{Extracting colon and polyp patches from real videos}}: this data collection consists of 97 cases, documented with a colonoscopy video per case, captured at the Hospital Universitario Nacional de Colombia between September 2018 and February 2019. Specifically, 59 lesions were observed in 50 cases reported with at least a superficial neoplastic lesion, and 47 cases were found to have no lesion. 100 squared patches were extracted from the colon wall with different sizes ($263 \pm 121$ pixels), and 59 squared patches from polyp lesions ($95 \pm 14$ pixels).
    
    \item \textbf{\textit{Qualitative evaluation of the SfSNet}}: the ETIS-Larib dataset collected 196 polyp frames with a resolution of $1\,225 \times 966$. The authors reported that these frames were extracted from 34 colonoscopy videos with 44 different polyps, without temporal continuity. Each of these images contains at least one polyp, delineated by a ground truth mask \citep{silva2014toward}.
\end{itemize}


\section{EVALUATION AND RESULTS}
\label{sec:result_sec}

Section~\ref{subsec:eval_synthetic_colon} shows some examples of the five levels of detail and illustrates the depth maps of each. Next, the depth estimation method is evaluated with three different tests. First, the two learning strategies (traditional {\em and} curriculum) estimate the colon depth in the synthetic database (see Section~\ref{subsec:eval_synth}), each of them implemented with three network architectures (DenseNet169, ResNet50, and EfficientNetB0). Second, these trained models were tested in a public real colonoscopy database (see Section~\ref{subsec:eval_real}), but only qualitative results are shown. The best configuration of the method is the one with better performance metrics along with the best visual depth estimation. Performance of this  configuration was compared with five state-of-the-art works (see Section~\ref{subsec:eval_comparison_methods}) as follows: (a) the method was assessed quantitatively using a public synthetic database with depth annotations, and (b) a visual evaluation of the depth estimation of real images extracted from these works.


\subsection{Synthetic Colonoscopy dataset }
\label{subsec:eval_synthetic_colon}

\begin{figure*}[ht!]
\centering
\includegraphics[scale=0.8]{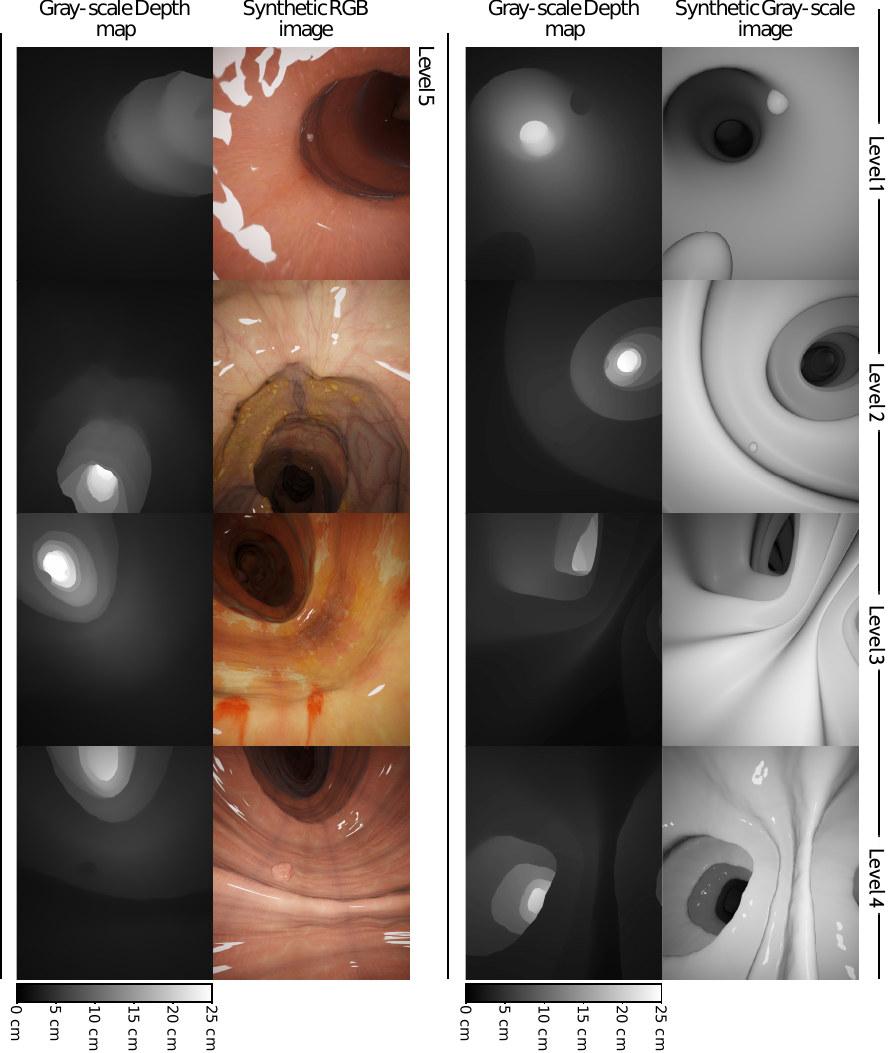}
\caption{Frames with depth annotations for each level of the synthetic video collection. Levels 1 to 4 are shown in the top panel, where synthetic frames are in the first row along with their corresponding depth maps in second row. Bottom panel shows four pictures corresponding to level 5. A gray-scale bar at the right of depth map rows shows the corresponding depth values in centimeters.}
\label{fig:synthetic_collection_samples}
\end{figure*}

A subset of videos was generated for each level of detail. Figure~\ref{fig:synthetic_collection_samples} shows frames from each subset, illustrating integration of details for each level. In Level 1 (top panel - first column) a very basic colon includes a coarse polyp, and the resultant depth map shows soft depth gradients, except around the polyp where larger gradients are observed. Level 2 (top panel - second column) incorporates haustral folds, producing more complex patterns of gradients, and some regions occluded by folds. Level 3 (top panel - third column) shows a deformed view of the colon with a triangular lumen shape. Likewise, Level 4 (top panel - fourth column) integrates the inner irregularities and specular reflections, producing a much more realistic surface, but with no major change in the depth map since the resultant gradients are very small. Furthermore, specular reflections (white regions) provide information about the surface orientation with respect to the light source, and modifications of depth maps are minor. The bottom panel of Figure~\ref{fig:synthetic_collection_samples} shows four samples of Level 5, illustrating how real the generated views look. In the first column, huge specular reflections are observed when the colonoscope tip is close to the colon wall. This particular view looks similar to the ascending colon, without pronounced haustral folds. The second column shows a frame with clear-cut blood vessels and feces. Observe the two protruding structures from the colon wall which look like polyps, one near to the center and the other to the bottom-right, but they are actually typical fold wrinkles. In the third column, red regions simulate the mucosa bleeding and three polyps are located at the bottom of the cavity, clearly observable in the depth map. Finally, in the fourth column, a small sessile polyp, hidden by its proximity to a haustral fold, shows a tissue appearance very similar to the colon wall. This set of images reproduces scenes with both structures very near to the camera (darker regions in depth map) or far from it (whiter regions), according to the gray-scale at the right. In total, 47 synthetic videos were rendered as follows: Levels one to four have 5 videos each, and Level five has 27 videos. The complete database used in this work is released for public access in \url{http://cimalab.unal.edu.co/?lang=en&mod=program&id=28}. 

\subsection{Training and evaluation with the Synthetic Colonoscopy dataset}
\label{subsec:eval_synth}

Performance of the depth estimation method with synthetic data is assessed by comparing the predicted depth $\hat{y}$ with the ground truth depth $y$ for each frame. 


\subsubsection{Experimental setup}
\label{subsubsec:global_experimental_setup}

\paragraph{\textbf{- Dataset:}} The Synthetic Colonoscopy dataset was split as shown in Table \ref{table:table_dataset_results}. For CL from Levels 1 to 4, 80\% of the dataset was used to train and 20\% for validation. Given that  CL versus TL comparison was performed until Level 5, to guarantee a similar variance for both strategies, the same quantity of data was for training (55\%), validation (15\%), and testing (30\%).

\begin{table}[ht]
\centering
\small
\renewcommand{\arraystretch}{1}
\begin{tabular}{ccccccc}
\toprule
 \multirow{2}{*}{\textbf{\begin{tabular}[c]{@{}c@{}}Learning\\ strategy \end{tabular}}} & \multirow{2}{*}{\textbf{Partition}} & \multicolumn{5}{c}{\textbf{Level of detail}} \\ \cline{3-7}
 &  & \textbf{L1} & \textbf{L2} & \textbf{L3} & \textbf{L4} & \textbf{L5}
\\ \cline{1-7}
 & {\textit{Train}} & { -} & { -} & { -} & { -} & {\textbf{15}} \\ 
 \multirow{-2}{*}{\begin{tabular}[c]{@{}c@{}}\textbf{Traditional} \end{tabular}} & { \textit{Validation}} & { -} & { -} & { -} & { -} & {\textbf{4}} 
\\ \cline{1-7}
 & { \textit{Train}} & { 4} & { 4} & { 4} & { 4} & { \textbf{15}}  \\ 
 \multirow{-2}{*}{\textbf{\begin{tabular}[c]{@{}c@{}}Curriculum \end{tabular}}} & { \textit{Validation}} & { 1} & { 1} & { 1} & { 1} & {\textbf{ 4}}
\\ \cline{1-7}
 & \textit{Test} & { -} & { -} & { -} & { -} & \textbf{ 8} \\
\toprule
\end{tabular}
\caption{Distribution of synthetic video collection for traditional and curriculum learning strategies.}
\label{table:table_dataset_results}
\end{table}


\paragraph{\textbf{- Network architectures:}} Three networks with different architectures were assessed. These architectures are further described below.

\begin{itemize}
    \item \textit{ResNet50:} This consists of 50 layers with 25.5 million parameters. Particularly, ResNet family incorporates the so-called residual blocks with shortcut connections, which mitigate the vanishing gradient limitation \citep{he2016deep}.
    \item \textit{DenseNet169:} A network built with 169 layers corresponding to 14.3 million parameters. Each layer  is fed from previous layers, i.e. each layer concatenates feature maps from all preceding layers \citep{huang2017}.
    \item \textit{EfficientNetB0:} an adjusted ResNet architecture which was scaled in terms of the number of layers and channels. EfficientNetB0 is a deep network with 237 layers but only 5.3 million parameters \citep{tan2019}.
\end{itemize}


\paragraph{\textbf{- Network optimization:}} The hyper-parameter optimization avoids overfitting or underfitting, problems frequently described with synthetic databases, that only partially represent real data. A Bayesian optimizer finds out an optimal combination of 5 hyper-parameters during two stages of 75 trials: in the first stage, the batch size, initial learning rate and weight decay are optimized. In the second stage, two parameters are optimized, (a) the factor by which the learning rate is decreased after a certain number of epochs without network improvement, and (b) the patience parameter, i.e., the number of epochs without improvement needed to decrease the learning rate. This process separately applies for each combination of the two learning strategies, and the three network architectures. In case of CL, this optimization process adapts to the data complexity at each level. Further details of this process are presented in Appendix~\ref{appendix_network_optimization}


\paragraph{\textbf{- Evaluation metrics:}} Two typical pixel-wise evaluation metrics are used to compare the estimated depth image $\hat{y}$ and the ground truth $y$, namely Root mean squared error (RMSE) and Threshold accuracy (ThAcc), described below:

\begin{itemize}
\item \textit{Root mean squared error (RMSE), in centimeters}:  
$$
\displaystyle \sqrt{\frac{1}{n}\sum_{p=1}^{n}(y_{p}-\hat{y}_{p})^2}
$$

\item \textit{Threshold accuracy (ThAcc), in \%:}
$$
\frac{100 \times \left| \left\{p ; \max \left( \frac{y_{p}}{\hat{y}_{p}}, \frac{\hat{y}_{p}}{y_{p}} \right) < \delta \right\} \right|}{n}
$$
Where $\delta$ is a decision threshold of the $y$ and $\hat{y}$ ratio, herein set to 1.25 as the strictest used in
the literature. For both metrics, $p$ is the pixel index, $n$ is the total number of pixels, and vertical bars $|S|$ denote the cardinality of the set $S$ in evaluation.
\end{itemize}


\paragraph{\textbf{- Three-dimensional reconstruction from depth map:}} An RGB image together with the corresponding depth map allows to project from the $2d$ image coordinate system to the $3d$ world coordinate system. This $3d$ reconstruction is used to evaluate visually the depth estimation. Therefore, a pixel in $(u,v)$ with depth $d$ is mapped to the $(x,y,z)$ three-dimensional world coordinates ($cm$) using a transformation matrix composed of the intrinsic parameters of the virtual or real camera, as follows:
$$
\begin{pmatrix} x & y &z \end{pmatrix}^T = d K^{-1} \begin{pmatrix} u & v & 1 \end{pmatrix} ^T
$$
with:
$$
K^{-1} = \begin{pmatrix}
1/f & 0   & -c_x/f \\ 
0   & 1/f & -c_y/f\\ 
0   & 0   & 1\\ 
\end{pmatrix} 
$$

Where $f$, the focal distance, is set to $0.1755 \: cm$ (448.13 pixels) corresponding to an endoscope wide angle of $110^{\circ}$, and a focal axis point located in the image center, $(c_x,c_y) = (0,0)$. 

The result of this transformation is a cloud of points forming a partial three-dimensional reconstruction of the colon wall. Additionally to the $3d$ reconstruction, 2.5d surface visualization is presented, as \citet{mahmood2018} and \citet{rau2019} provided in their work by plotting the depth $z$ as function of $(u,v)$.


\subsubsection{Ablation study of loss function $L_{SfSNet}$}
\label{subsubsec:loss_ablation_study}

The impact of terms $L_{z}$, $L_{e}$ and $L_{c}$ of the loss function $L_{SfSNet}$, presented in Section~\ref{subsubsec:custom_loss}, was evaluated with 4 experiments corresponding to (1) $L_{z}$ alone as the baseline, (2) $L_{e}$ and (3) $L_{c}$ combined with $L_{z}$, and (4) the three terms combined with the weights selected in this work. For these experiments, EfficientNetB0 architecture following a TL strategy was used. The results were reported in Table~\ref{table:results_ablation_loss}

\begin{table}[h!]
\centering
\renewcommand{\arraystretch}{1.1}
\begin{tabular}{llcc}
\toprule
No.& \multirow{2}{*}{\textbf{Experiment $L_{SfSNet} =$}} & \multicolumn{2}{c}{\textbf{Metrics}} \\
\cline{3-4}
& & \textbf{ThAcc (\%)} & \textbf{RMSE} \\
\cline{1-4}
1 & $L_{z}$ & 89.52 & 0.607 \\
2 & $0.5*L_{z} + 0.5*L_{e}$ & 94.13 & 0.570 \\
3 & $0.5*L_{z} + 0.5*L_{c}$ & 94.61 & 0.554 \\
4 & $0.1*L_{z} + 0.3*L_{e} + 0.6*L_{c}$ & 96.13 & 0.490 \\
\toprule
\end{tabular}
\caption{Evaluation metrics computed with respect to the ground truth obtained by the proposed method with different configurations of loss function terms.}
\label{table:results_ablation_loss}
\end{table}

Table~\ref{table:results_ablation_loss} shows 
that
combination of $L_{e}$ and $L_{c}$ terms along with $L_{z}$ in the cost function $L_{SfSNet}$ (Exp. 4: 96.13\% of ThAcc and $0.490 \: cm$ of RMSE) exhibits a performance improvement with respect to the $L_{z}$ term alone (Exp. 1: 89.52\% of ThAcc and $0.607 \: cm$ of RMSE). Likewise, experiments 2 and 3 evidence that separated $L_{e}$ and $L_{c}$ evaluation show improvement, but less than combination of the three terms.


\subsubsection{Quantitative results}
\label{subsubsec:quantitative_synthetic}

Table~\ref{table:results_synthetic} presents quantitative results with the testing set of Level 5. In general, CL overperforms TL strategy, likely indicating it is worthy to model complexity and progressively introduce it. Specifically, in case of ResNet50, CL is larger in ThAcc and RMSE, with $92.55\%$ and $0.496$ respectively, while $90.46\%$ of ThAcc and $0.563$ of RMSE were observed for TL. For DenseNet169, network performance was much better for CL with $94.98\%$ of ThAcc and $0.508$ of RMSE, in contrast to TL, which obtained $74.32\%$ of ThAcc and $0.715$ of RMSE. Finally, the EfficientNetB0, the smallest architecture, following a TL strategy, reached a $96.13\%$ of ThAcc, slightly better than CL which obtained $95.65\%$ of the same metric. However, RMSE remains better for CL, i.e., TL obtained $0.490$ and CL reached $0.451$. This network achieved the best  ThAcc and RMSE performance when comparing to other nets. 

An extended analysis of the validation and training losses for both training strategies is presented in Appendix~\ref{subappendix_eval_synthetic_quanti}. Additionally, for CL, Levels 1 to 4 were challenged using the testing set of Level 5 to analyze progression at each level, which is also presented in Appendix~\ref{subappendix_eval_synthetic_quanti}.

\begin{table}[htb]
\centering
\renewcommand{\arraystretch}{1.1}
\begin{tabular}{llcc}
\toprule
\multirow{2}{*}{\textbf{Network}} & \multirow{2}{*}{\textbf{Metrics}} & \multicolumn{2}{c}{\textbf{Learning strategies}} \\
\cline{3-4}
& & \textbf{Curriculum} & \textbf{Traditional} \\
\cline{1-4}

\multirow{2}{*}{\textbf{ResNet50}}
& \textbf{ThAcc (\%)} & \textbf{92.55} &  90.46 \\
& \textbf{RMSE} & \textbf{0.496} & 0.563 \\
\cline{1-4}

\multirow{2}{*}{\textbf{DenseNet169}}
& \textbf{ThAcc (\%)} & \textbf{94.98} & 74.32 \\
& \textbf{RMSE} & \textbf{0.508} & 0.715 \\
\cline{1-4}

\multirow{2}{*}{\textbf{EfficientNetB0}}
& \textbf{ThAcc (\%)} & 95.65 & \textbf{96.13} \\
& \textbf{RMSE} & \textbf{0.451} & 0.490 \\
\toprule
\end{tabular}
\caption{Evaluation metrics computed between the ground truth and the predicted depth obtained by the proposed method using the testing set of Level 5. Traditional and curriculum learning were tested (columns), with three networks (rows).}
\label{table:results_synthetic}
\end{table}


\subsubsection{Qualitative results}
\label{subsubsec:qualitative_synthetic}

Provided EfficientNetB0 was the best after results presented in section~\ref{subsubsec:quantitative_synthetic}, this network was the one used for the next set of experiments: comparing depth estimations of the final CL and TL models and reconstructions $3d$ from the depth estimation maps. 

\begin{figure*}[ht!]
\centering
\includegraphics[scale=1]{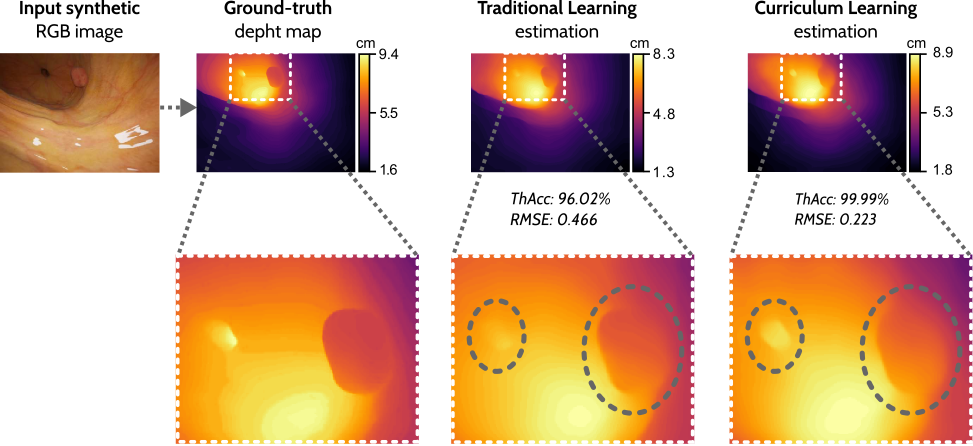}
\caption{Depth map of a synthetic image is compared with estimations from the two learning strategies on EfficientNetB0. A zoomed view of the region of interest shows the estimation of a medium-size polyp and caecum-appendix aperture.}
\label{fig:qualitative_synthetic_multiple}
\end{figure*}

The CL against TL comparison is shown in Figure~\ref{fig:qualitative_synthetic_multiple}. CL approximated better than TL the shape of a medium-size polyp and caecum-appendix aperture, in contrast to TL which only captured the polyp. This result is coherent with the performance metrics: $99.99\%$ against $96.02\%$ of ThAcc were respectively obtained for CL and TL, while their respective RMSE corresponded to $0.223 \: cm$ and $0.466 \: cm$. Furthermore, while CL reached any depth, TL estimated up to $8.3$ cm and nearby regions showed that CL also estimated more accurately the corresponding depth, i.e., CL estimates start from $1.3$ cm and TL ones from $1.8 \: cm$. From now on, all experiments will be run with EfficientNetB0 and CL: referred as SfSNet for simplicity. 

\begin{figure*}[ht!]
\centering
\includegraphics[scale=0.8]{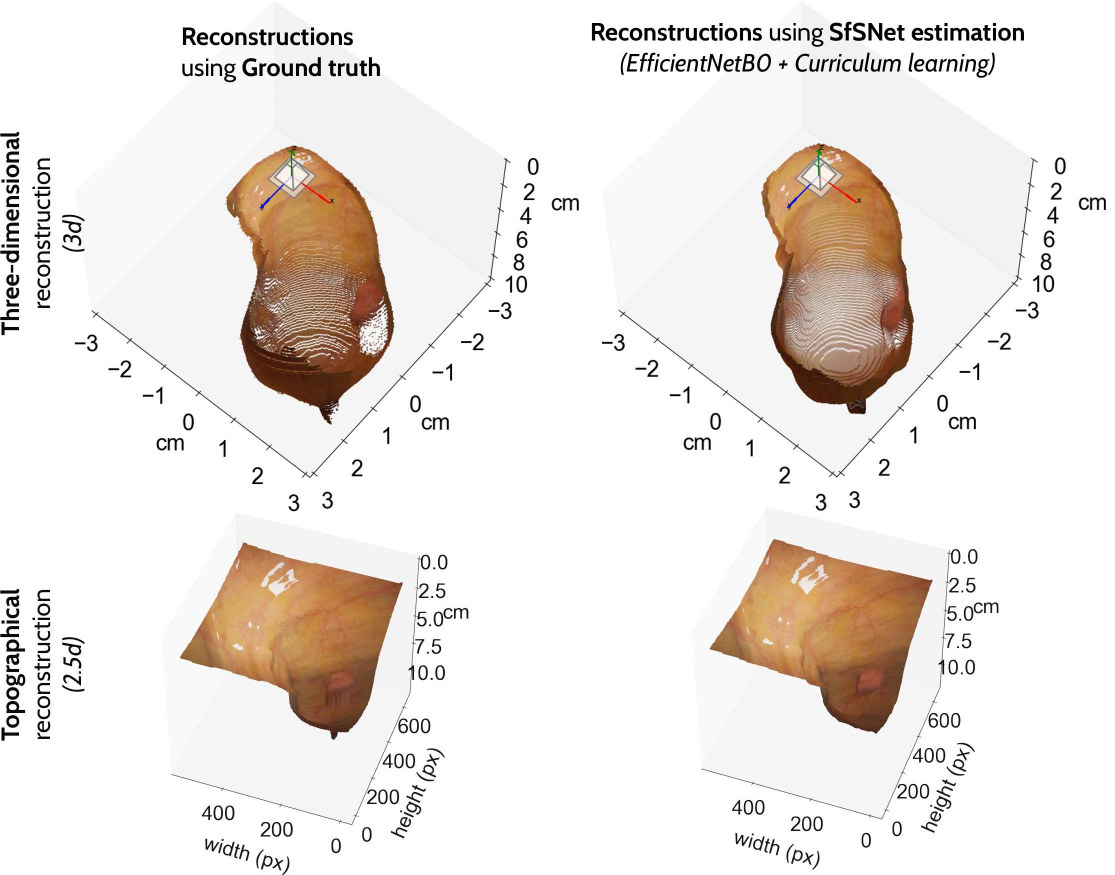}
\caption{Three-dimensional reconstructions (first row) and topographical surfaces (second row) from ground truth depth map (first column) and SfSNet (second row) of the synthetic image tested in figure~\ref{fig:qualitative_synthetic_multiple}. Each $3d$ reconstruction includes a small pyramid that represents the virtual camera pose.}
\label{fig:3d_projections_synthetic}
\end{figure*}

$3d$ and $2.5d$ (topographical) reconstructions of the synthetic image previously tested are presented in Figure~\ref{fig:3d_projections_synthetic}. Reconstructions by SfSNet are close to the ground truth, the caecum with appendiceal orifice and a polyp of medium size ($\sim 1\: cm$) are perfectly observable. Even under poor illumination conditions and large specular reflections in the farthest regions, the method reliably estimates the colon geometry.

In addition, Levels 1 to 4 models of CL were challenged using a synthetic image from the testing set of Level 5 to analyze progression at each level in Appendix~\ref{subappendix_eval_synthetic_quali}. 

\subsection{Evaluation of the depth estimation with Real colonoscopy datasets}
\label{subsec:eval_real}

\begin{figure*}[ht!]
\centering
\includegraphics[scale=0.7]{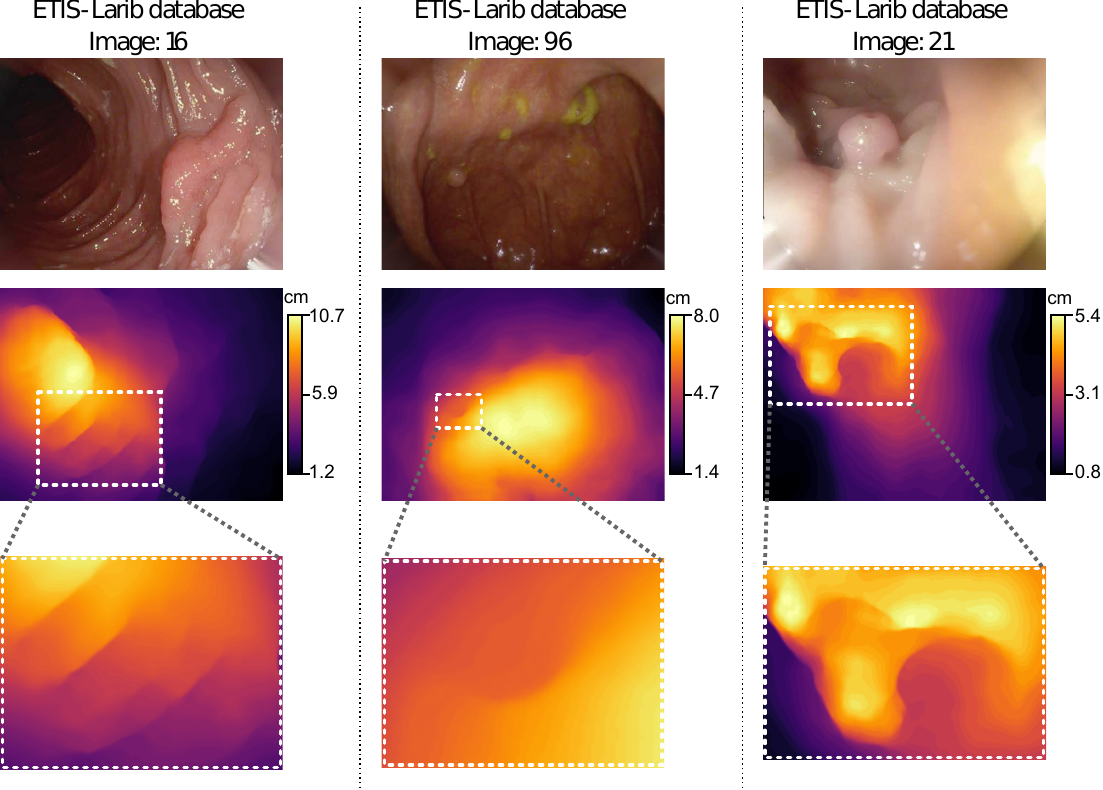}
\caption{Depth estimations of Curriculum learning strategy with EfficientNetB0 (SfSNet) on three real images. The first row shows the real RGB images, second row the SfSNet estimations and third row a zoomed region from the second row.}
\label{fig:qualitative_real_multiple}
\end{figure*}

This section is devoted to qualitatively assess the estimated depth in a real colonoscopy collection, the ETIS-Larib dataset \citep{silva2014toward}, by a SfSNet model trained only with synthetic data (EfficientNetB0 plus CL strategy). Evaluation focused mainly on how the method captured polyp shapes with different sizes, as well as nearby folds at different distances. In addition, $3d$ and $2.5d$ reconstructions were coherent with what is observed in an image: three examples are shown in Figure~\ref{fig:qualitative_real_multiple}. In the left column, reconstruction generates a mid-range depth and a sessile polyp (0-Is according to Paris classification). In this case, depth of the small folds in the deepest regions was very well estimated, as observed in the zoomed region. In the center column, the method was able to reliably estimate the boundary of a tiny polyp and residual stools in a mid-range depth scenario. The right column shows a short-range depth corresponding to an under-distended colon wall and a polyp, forming a complex surface that was not modeled by the synthetic database. However, the method successfully captured this geometry and also delineated accurately the polyp boundaries and small folds, even in regions that are shadowed by a large fold. In general, depth ranges presented in color bars are coherent with apparent distances between these structures and the camera, i.e., left image shows a depth range estimated around $11 \: cm$, $8 \: cm$ for the image with mid-range depth (center column), and $5 \: cm$ for the image with short-range depth (right column). 

\begin{figure*}[ht]
\centering
\includegraphics[scale=0.9]{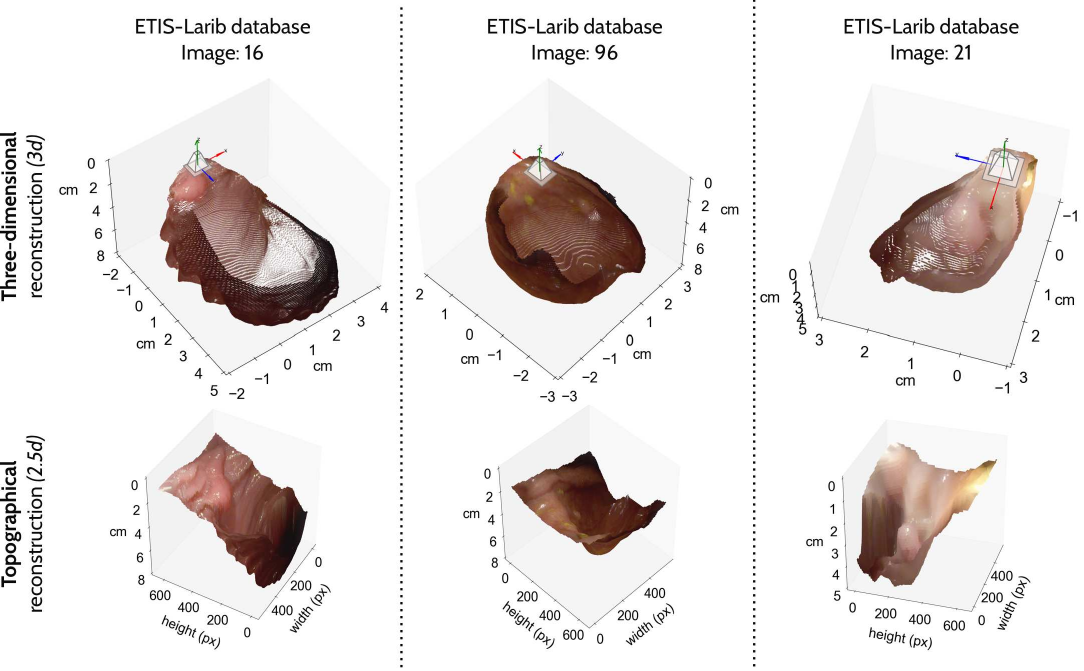}
\caption{Three-dimensional reconstructions (first row) and topographical surfaces (second row) from estimated depth maps by SfSNet in three real images. The columns correspond to each image tested in figure \ref{fig:qualitative_real_multiple}. Each $3d$ reconstruction includes a small pyramid that represents the virtual camera pose.
}
\label{fig:3d_projections_real_own}
\end{figure*}

In the three images, $3d$ reconstructions presented in Figure~\ref{fig:3d_projections_real_own} shows the geometry of each colon structure, shape of folds in the left column, different size polyps clearly highlighted as protrusions in the mid column and a reliable complex geometry reconstruction in a short-range scene in the right column. 

\subsection{Comparison with state-of-the-art methods}
\label{subsec:eval_comparison_methods}

Performance of the proposed method trained with synthetic data was twofold assessed with state-of-the-art methods. The method was first challenged by a public synthetic database (collection of \citet{rau2019}), and then with real colonoscopy images used in five different works \citep{rau2019,mahmood2018,freedman2020,itoh2021,mathew2020}. It is worth mentioning that none of these works released the code, making it difficult to quantitatively compare with them. Furthermore, several implementation details are not provided in the manuscripts, which limits
the possibility of fair comparison. Therefore, comparison is qualitative, as some authors have done in their publications \citep{cheng2021,mahmood2018}. 

\subsubsection{Evaluation in a public synthetic database}
\label{subsubsec:evaluation_public_synthetic_database}

This experiment aims to quantitatively assess the SfSNet performance with synthetic data different from those generated in this work, i.e., a public and ready-to-use collection released by \citet{rau2019}. Briefly, the collection of \citet{rau2019} is composed of 45 virtual colonoscopy videos generated from a single CT-based colon model in a licensed $3d$ computer graphics platform (Unity), varying color, intensity, reflectivity, surface irregularities, and aperture angle of the source, all of them integrated to a virtual camera following a random path. In total, 16\,000 RGB images with depth maps normalized from 0 to $20 \:cm$ were released. Estimated depth maps by SfSNet were two-fold evaluated by computing RMSE: either (a)  with all pixels of the video, reporting the mean-RMSE for the entire set of videos or (b) grouping all pixels in 18 depth ranges.

The complete collection (16\,000 images) was tested by SfSNet obtaining a mean-RMSE of $0.967\pm 0.169 \: cm$. Notice no training whatsoever was performed with this database, only testing, and interestingly a similar figure was obtained when different subsets of the database were assessed (30 \% and 50 \%), strongly supporting the generalizability of the presented method. \citet{rau2019} in contrast reported $0.175 \: cm$ of mean-RMSE using only a $30\%$ of database (approximately 4\,500 images) since the remaining 70 \% was used to train. This result can not be fairly compared with the RMSE obtained by the SfSNet, yet this is close, although it was not trained with the same set of data.

\begin{figure}[h!]
\centering
\includegraphics[scale=0.9]{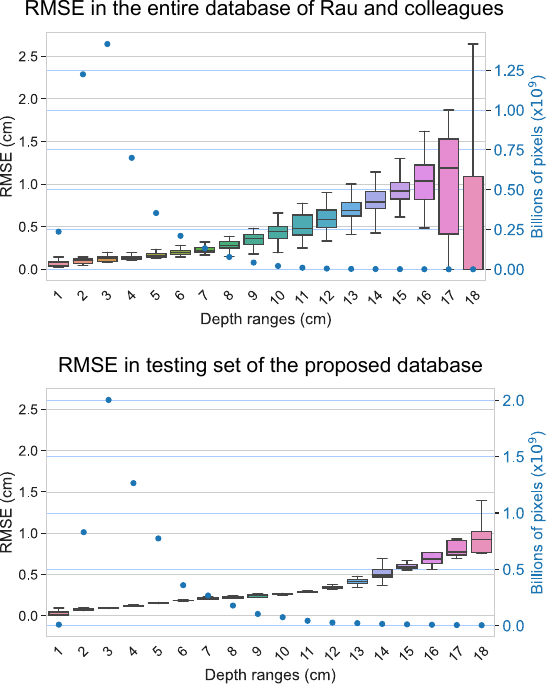}
\caption{Root mean square error for eighteen depth ranges in the CT-based database of \citet{rau2019} (top) and in the synthetic database herein presented (bottom). The blue dot curve represents the number of pixels in each depth range. 
}
\label{fig:depth_range_analysis}
\end{figure}

Performance of SfSNet was also evaluated in 18 depth ranges since the distance to the camera influences the result of the estimation. As shown in figure~\ref{fig:depth_range_analysis}(top), which displays in box plots the RMSE distribution for each range, SfSNet achieved a better performance in closer regions (from $0$ to $10 \: cm$) with a RMSE less than $0.5 \: cm$ and a quite small variance. In ranges from $10$ to $18\:cm$, RMSE did not exceed $1.25 \: cm$, a difference probably explained by light source variations. For comparison, this analysis was also carried out in the database herein constructed using a testing set corresponding to a 30 \%  (21\,600 frames) of the total of videos. As shown in figure~\ref{fig:depth_range_analysis} (bottom), the SfSNet reached better performance in closer regions, with RMSE less than $0.3 \: cm$, while  SfSNet did not exceed $0.8 \: cm$ in the Rau database, with a much smaller deviation. 

\subsubsection{Qualitative evaluation with real colonoscopy images}
\label{subsec:evaluation_public_synthetic_database}

\begin{figure*}[ht]
\centering
\includegraphics[scale=0.9]{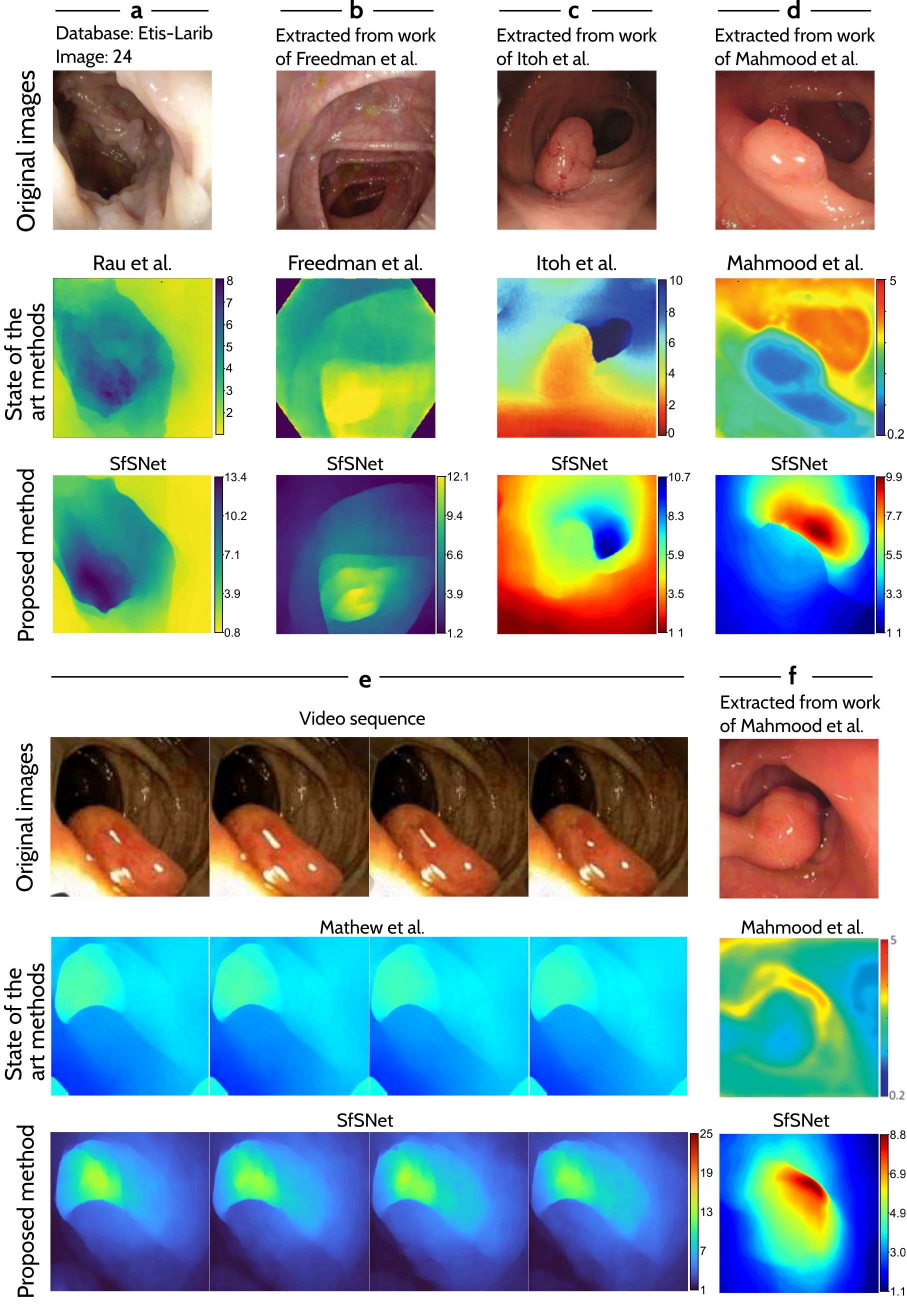}
\caption{Comparison between SfSNet versus five state-of-the-art works estimating depth in real colonoscopy images. Each column is composed of three rows: RGB image in the first row, estimated depth maps by one of the five methodologies in the second row, and estimated depth maps by SfSNet in the third row.}
\label{fig:state_of_art_selected}
\end{figure*}

A qualitative comparison of four real colonoscopy frames is presented in figure \ref{fig:state_of_art_selected} between SfSNet against four state-of-the-art works, which are \citet{rau2019}, \citet{mahmood2018}, \citet{freedman2020}, \citet{itoh2021}. SfSNet reliably estimated shape of folds and polyps in images \textit{a} and \textit{b}, capturing very well borders at different depths, compared with blurred predictions by \citet{rau2019}, and \citet{freedman2020}. In images \textit{c} and \textit{d}, SfSNet generated coherent depth maps capturing remarkably well the cylindrical shape observed in these perspectives of colon, and without overestimating depth around polyps or folds. Although estimations obtained by \citet{itoh2021} in image \textit{c} and \citet{mahmood2018} in image \textit{d} were reasonably good, those methods produced flat regions orthogonal to camera axis instead of the cylindrical shape. Apparently, the method of \citet{itoh2021} is prone to failure in shadows produced by polyps or folds, generating overestimated gradients of depth. All previous images belong to ETIS-Larib \citep{silva2014toward} and CVC-ColonDB \citep{bernal2015} databases, but in most cases, images were extracted directly from the manuscripts since the authors did not provide details to identify them. 

Also, SfSNet was challenged in a sequence of temporal frames (\textit{e} in figure \ref{fig:state_of_art_selected}) presented by \citet{mathew2020} to illustrate consistency of depth estimations along time. Even though SfSNet did not include a temporal regularization, the obtained temporal sequence was also coherent since depth maps look very similar among frames, preserving soft gradients in colon wall and strong gradients around polyp. 

\begin{figure*}[ht]
\centering
\includegraphics[scale=0.9]{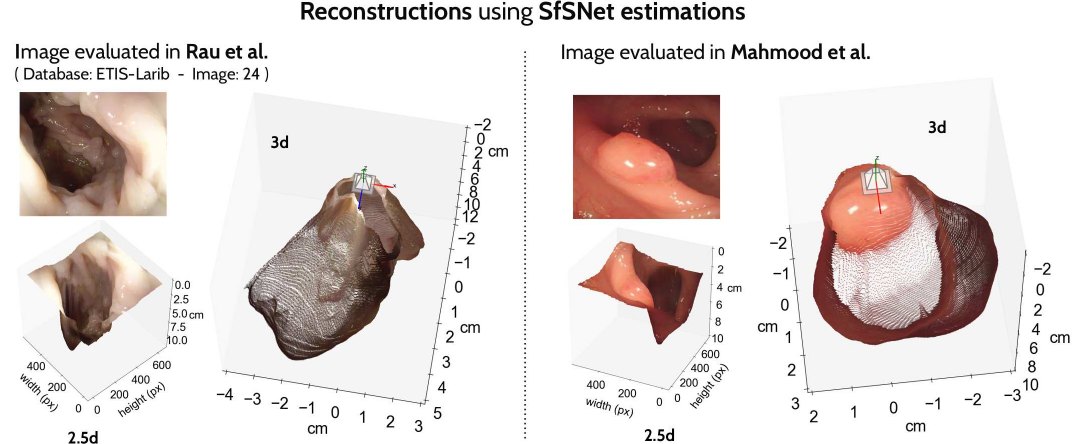}
\caption{Two examples of three-dimensional reconstruction obtained from single depth maps estimated in real colonoscopy images used in the state-of-the-art works. Each figure is composed of a RGB image, depth visualized as a topographical surface, and three-dimensional reconstruction.}
\label{fig:state_of_art_selected_retro}
\end{figure*}

Most papers of the state of the art show no reconstructions with the estimated depth. Here, $3d$ and $2.5d$ reconstructions of two images from previous comparison are presented in Figure~\ref{fig:state_of_art_selected_retro}.  In the left panel, a narrow colon perspective, with folds close to camera, is accurately estimated. In this particular case, $2.5d$ is useful to show reliable estimation in the deepest part of this complex geometry. Finally, the polyp in the right panel shows a protruded shape with a reconstruction discontinuity.

Finally, some limitations of SfSNet are shown in image \textit{f} of figure \ref{fig:state_of_art_selected}, where borders of a huge pedunculated polyp were wrongly estimated, likely because this size of lesions was not included in the training set of synthetic database, and basically since they pose no recognition problem to the clinician. However, the nearest and farthest regions were much better predicted by SfSNet than \citet{mahmood2018}.

\section{Discussion}\label{sec:discussion}

This work has introduced an automatic strategy to estimate the depth map of single frames from monocular colonoscopy videos. In comparison with the state of the art, this strategy presents substantial advantages: (a) the use of raw RGB images previously preprocessed, (b) a simple network architecture driven by a customized loss function specially adapted to the colonoscopy workflow, and (c) construction of a fully-synthetic database available for open use (\url{http://cimalab.unal.edu.co/?lang=en&mod=program&id=28}). Overall, they are all complementary steps for the construction of accurate depth maps which are needed to describe and characterize colon lesions.

SfSNet preprocessing includes filling out corner black regions, typical of colonoscopy, and standard image resizing to feed the network, making SfSNet more robust since it is trained with typical patient-specific textures and specular highlights. In contrast, \citet{mathew2020}, \citet{mahmood2018}, and \citet{itoh2021} performed intensive preprocessing to map raw images' appearance to a virtual colonoscopy environment at a high computational cost, using visual transformers \citep{mahmood2018} or Generative Adversarial Networks (GAN) \citep{mathew2020,itoh2021}. These authors justified this preprocessing since their approaches were too sensitive to noise and variability from real images, e.g., depth on specularities or vessels was wrongly estimated. A much simpler and robust supervised solution is herein presented, composed of an encoder-decoder layout, specifically an EfficientNetB0 that estimates depth maps using customized colonoscopy features, essentially: (i) a customized loss function focused on estimating depth of curved shapes, like polyps and haustral folds; and (ii) a fully-synthetic database which can be easily enriched with different appearances. Unlike this approach, complex architectures were presented by \citet{rau2019}, \citet{mathew2020}, and \citet{itoh2021}, who implemented unsupervised GANs versions, i.e., a CycleGAN, composed of two GANs (four networks, two discriminators and two generators), or \citet{freedman2020} who implemented an unsupervised two-fold network. Of course, unsupervised approaches are the natural choice for a problem with no annotated data, an advantage with respect to SfSNet, but this choice involves exponential complexity with larger and larger training datasets, which implies (i) great difficulty to find an optimum with several networks connected \citep{porkodi2022,herr2021}, and (ii) a higher computational cost, limiting the clinical application. In particular, the optimum search is not a minor issue with these networks, for instance, \citet{rau2019} and \citet{freedman2020} reported the blurred estimations shown in section \ref{subsubsec:evaluation_public_synthetic_database}, while SfSNet properly estimated structural borders like haustral folds or polyps for a wide range of depths. Interestingly, some studies \citep{terada2022,liu2021} have already claimed that GAN-based approaches are limited at recovering fine-grained details, which in colonoscopy would mean to misestimate small polyps, probably the most important structure to deal with. Furthermore, Itoh \citet{itoh2021} reported $3d$ depth maps which do not match the cylindrical colon shape, while SfSNet always captured this shape as shown in Figures \ref{fig:3d_projections_real_own} and \ref{fig:3d_projections_synthetic}, which could indicate that depth colon estimation needs a certain amount of supervised training, especially to learn fine-grained details, such as small polyps.

The synthetic collection with depth annotations herein released is probably one of the most realistic in literature, e.g., the models of \citet{rau2019} or \citet{itoh2021} were built with very few textures and colors and without specular reflections, and in case of the collection of \citet{freedman2020}, it incorporates a realistic appearance in their models but no specular reflections. Overall, construction of colon and polyp shapes was herein fully controlled and largely automated, generating a set of 43 unique colon models and $\sim$ 430 polyps, a significant advantage with respect to CT-based databases, since generating new models with unique folds and polyps is a time-consuming manual task, for this reason, the collection of \citet{rau2019} used only a CT-based model from a single subject, while in the work of \citet{itoh2021} adapted only 9 models from different subjects. Even though CT-based collections capture real colon geometry, they have several drawbacks, namely: (i) low spatial resolution during in-vivo acquisitions which limits reconstruction of tiny polyps, (ii) operator-demanding task since each slide of a volume is manually segmented to produce the colon mesh, and (iii) post-processing is required to filter out artifacts from CT acquisitions \citep{mahmood2018}. Finally, it is noteworthy to mention that the collection herein introduced is ready to use, unlike the environment of data generation presented by Incetan \citep{incetan2021} which requires prior knowledge of several tools to render a video, or the synthetic videos published by Freedman \citep{freedman2020}, which do not contain depth annotations.

The method herein described has some limitations at estimating depth, mainly in atypical shapes or appearances not included in training set, e.g., huge polyps (see section \ref{subsec:eval_real}) which are not of great interest since they can be easily recognized. On the other hand, wrong estimations might occur when colon visualization is limited, e.g., when camera is occluded by bubbles or by a partially insuflated colon, or by a poor patient preparation, or when a considerable proportion of colon is covered by blood. Additionally, although SfSNet was trained with synthetic data that do not include different cameras (field of view or focal length) and light source configurations, the method produced coherent depth maps and $3d$ reconstructions for real databases captured with different optical characteristics. For $3d$ reconstructions, these drawbacks could be mitigated by knowing and using intrinsic real camera parameters during reconstruction.

Finally, the training strategies evaluated, CL and TL, obtained similar performance. Indeed, TL improved depth estimation by 7.4\% of ThAcc and $0.104 \: cm$ of RMSE for the three tested networks on average, and qualitatively, CL was able to capture more fine-grained details than TL. Although minor performance improvement was achieved by CL, this may represent crucial information like precancerous polyps. Likewise, CL was able to converge faster during training in two network architectures with respect to TL, which is consistent with what has been reported \citet{Surendranath2018} in indoor scenes. 

\section{Conclusions}\label{sec:conclusions}
 
This work presents a simple workflow with a minimal preprocessing step along with a network and a learning customized strategy producing consistent depth estimations in real images. The proposed method outperforms state-of-the-art works since this captured better fine-grained details and the cylindrical shape of colon. The synthetic database herein released includes much more colon and polyp variability in terms of appearance and shape with respect to current collections. Perspectives and future work in this method include: (1)~estimating camera pose to reconstruct the entire colon using techniques like Iterative Closest Point or bundle adjustment, (2)~modifying the loss function for enforcing temporal consistency of depth maps, (3)~developing a strategy to fine-tune the method for different configurations of camera and light sources, and (4)~further interpreting the effects of CL by analyzing the variations of activation maps among Levels. Finally, considering advantages of unsupervised models, the database released may contribute for the network, loss function, and learning strategy could be integrated to a GAN-based architecture.


\section*{Acknowledgments}
\begin{itemize}
    \item The methodological design of this work was carried out at ENSTA Paris, Institut Polytechnique de Paris, during the period from June to November 2019.
    \item This work was supported in part by the project with code 110192092354 and entitle ``Program for the Early Detection of Premalignant Lesions and Gastric Cancer in urban, rural and dispersed areas in the Department of Nari\~no'' of call No. 920 of 2022 of MinCiencias.
    \item This work was partially supported by project BPIN 2019000100060 ``Implementation of a Network for Research, Technological Development and Innovation in Digital Pathology (RedPat) supported by Industry 4.0 technologies'' from FCTeI of SGR resources, which was approved by OCAD of FCTeI and MinCiencias.
\end{itemize}

\section*{Declaration of competing interest}

The authors declare that they have no known competing financial interests or personal relationships that could have appeared to influence the work reported in this paper.

\section*{Appendices}
\appendix

\section{Network optimization}
\label{appendix_network_optimization}

The global configuration of the three architectures and the two stages of optimization are described below. This process was applied separately for each combination of the two learning strategies, and the three network architectures.

\begin{itemize}
\item \textit{Global configuration}:
    \begin{itemize}
    \item \textit{Loss component weights:}
    (See Equation~\ref{eq:global_loss})
    $w_{1}=0.1$, $w_{2}=0.3$ and $w_{3}=0.6$ according to the ablation study presented in Section~\ref{subsubsec:loss_ablation_study}
    \item \textit{Optimization algorithm}: Adam optimizer.
    \item \textit{Monitored metrics:} loss value in training and validation.
    \item \textit{Learning rate decay}: reduced by a factor $f$ when the validation loss does not improve for $n$ epochs.
    \item \textit{Gamma correction}: The value was manually tuned at $\gamma=0.66$
    \end{itemize}

\item \textit{First stage:} 60 trials of 10 epochs each assessed to find the best combination of the following hyper-parameters:
    \begin{itemize}
        \item \textit{Batch size:} Tested from 4 to 10.
        \item \textit{Initial learning rate:} Tested at $10^{k}$, for $k \in \{-6, -3\}$.
        \item \textit{Weight decay regularization:} This consists in adding the term $\lambda ||W||^2$ to the loss function, to bind the range of the weight values $W$. $\lambda$ was tested at $10^{k}$, for $k \in \{-6, -4\}$.
        \item \textit{Data partition:} 1/8 part of training and validation data sets was used at each trial.
        \item \textit{Weights initialization:} in case of CL, the network at first level is initialized from scratch, and for next levels, trained weights in previous level are used. For TL, weights are initialized from scratch.
    \end{itemize}
\item \textit{Second stage:} Using values of  three hyper-parameters previously found, 15 trials of 20 epochs were assessed to find the configuration of the learning rate decay strategy:
    \begin{itemize}
        \item \textit{Factor $f$:} Tested from 0.4 to 0.9.
        \item \textit{Patience $n$:} Tested from 4 to 10 epochs.
        \item \textit{Data partition:} 1/16 part of training and validation data sets were used in each trial.
    \end{itemize}
\end{itemize}

Finally, the 5 found hyper-parameters are used to train the networks. The loss training and validation histories are monitored at each level to manually select the epoch with the lowest stable validation loss, i.e., the training at a specific epoch is stopped when the training and validation loss values remain steady.

\section{Performance evaluation of training techniques in synthetic data}
\label{appendix_eval_synthetic}

Performance of the two learning strategies is compared quantitatively and qualitatively.

\subsection{Quantitative evaluation}
\label{subappendix_eval_synthetic_quanti}
Figure~\ref{fig:training_history} shows training and validation in terms of loss values for the three networks. The five CL training stages are observed in the different panels identified with the particular level: from one to four the CL losses are plotted in blue for training and red for validation. Since TL losses can be estimated only at Level 5, they are shown in green for training and purple for validation in the corresponding panel.

\setcounter{figure}{0}
\renewcommand{\thefigure}{A.\arabic{figure}}
\begin{figure*}[ht!]
\centering
\includegraphics[scale=0.8]{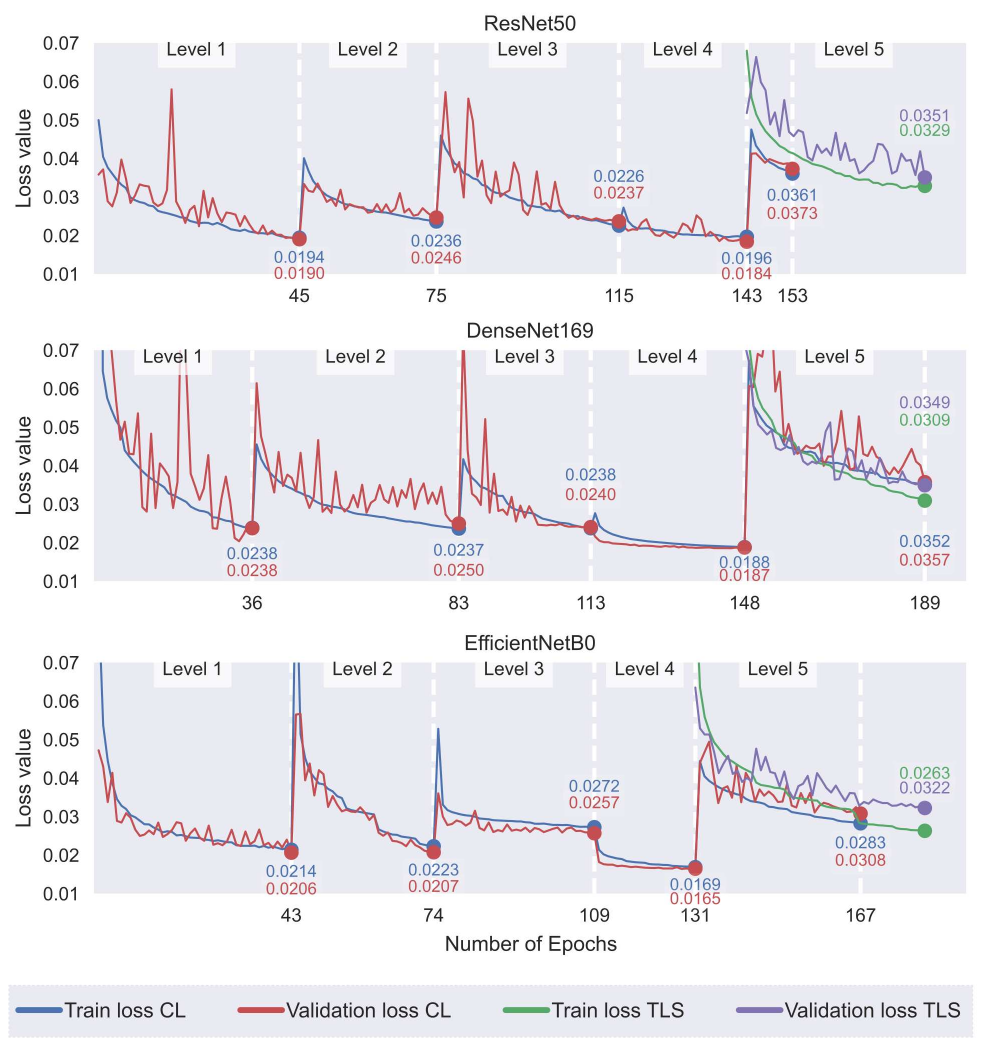}
\caption{Training and validation for the three nets: ResNet50 (top panel), DenseNet169 (middle panel) and EfficientNetB0 (bottom panel). The blue and red graphs correspond to the curriculum learning loss during the training and validation, respectively. Loss function of the traditional learning strategy is shown in green for training and purple for validation. Progressive curriculum learning levels are shown in panels from left to right.}
\label{fig:training_history}
\end{figure*}

In general, transition between two consecutive CL levels is characterized by a sudden increase of the loss function in both the training (blue) and validation (red) sets, since each level introduces information not known by the precedent neural models. The initial high value of the loss function rapidly falls down, quite regularly for the training set and irregularly for the validation set, and usually after waiting the loss stabilizes the number of epochs end up by being quite similar. Anyway, the number of epochs required for the next training set varies among architectures, in particular DenseNet169 requires more training time. Interestingly, the loss peak between Levels 3 to 4 is smaller than other transitions, likely because variability in Level 4 is much lower compared to the variability learned in the previous levels since specular reflections and surface irregularities are local phenomena. In fact, one can see these specular and irregular phenomena as the composition of much simpler spatial configurations compared to those learned in previous levels. More fundamentally, specularities with a controlled illumination actually help to recover shape, since the light source aligned with the camera's optical axis generates reflections only in surface regions whose normal is orthogonal to the focal plane.

In the TL case, the function loss in training (green graphs) and validation (purple graphs) shows a similar pattern for the three networks. Nevertheless, these losses are not so close at the final epoch, suggesting each network is approximating different parts of the Level 5 feature space. In contrast, CL based networks exhibit a better regularization and learning of Level 5 concepts. Additionally, the initial training and validation loss values were smaller for CL, at least for ResNet50 and EfficientNetB0, evidencing that initialization with pre-trained CL weights from Level 1 to 4 improves depth estimation during Level 5 training. 

Quantitative results with the testing set of Level 5 are presented in Table~\ref{table:results_synthetic}. Analyzing the final network performance and learning strategies, in case of ResNet50, although TL validation loss (last value of purple graph) is smaller than CL (last value of red graph), CL is larger in ThAcc and RMSE, with a $92.55\%$ and $0.496$ respectively, while $90.46\%$ of ThAcc and $0.563$ of RMSE were observed for TL. This network took advantage of using pre-trained  CL weights from Levels 1 to 4, i.e., CL took 10 epochs to reach a stable performance, while TL took 45 epochs. For DenseNet169, validation losses of both learning strategies are similar, but network performance was much better for CL with $94.98\%$ of ThAcc and $0.508$ of RMSE, in contrast to TL, which obtained $74.32\%$ of ThAcc and $0.715$ of RMSE. Finally, the EfficientNetB0, the smallest architecture, showed CL validation loss was better than TL strategy, because TL took more epochs to obtain a stable performance.  EfficientNetB0, following a TL strategy, reached a $96.13\%$ of ThAcc, a bit better than CL which obtained $95.65\%$ of the same metric. However, RMSE remains better for CL, i.e., TL obtained $0.490$ and CL reached $0.451$. This network achieved the best performance in ThAcc and RMSE compared to other nets.

In addition, for CL, Levels 1 to 4 were challenged using the testing set of Level 5 to analyze progression at each level. Still, incremental performance was not observed for all networks, e.g., for Levels 1 to 3, models had learned colon and polyp shapes at different complexity but they never learned the specular reflections and texture of Level 5. Likewise, performance in Level 4 decreased even more than in previous levels for all nets, probably because they confuse specular reflections (observed like white regions) learned in grayscale at Level 4, with color patterns close to white, that can be introduced in the tissue appearance at Level 5.

\setcounter{table}{0}
\renewcommand{\thetable}{B.\arabic{table}}
\begin{table*}[ht]
\centering
\renewcommand{\arraystretch}{1.2}
\begin{tabular}{P{2.5cm}P{2cm}P{1cm}P{1cm}P{1cm}P{1cm}P{2cm}P{2cm}}

\toprule
\multirow{3}{*}{\textbf{Network}} & \multirow{3}{*}{\textbf{Metrics}} & \multicolumn{6}{c}{\textbf{Learning strategies}} \\
\cline{3-8}
& & \multicolumn{5}{c}{\textbf{Curriculum} (Levels)} &\textbf{Traditional} \\
\cline{3-7}
& & \textbf{L1} & \textbf{L2} & \textbf{L3} & \textbf{L4} & \textbf{L5} & \textbf{L5} \\
\cline{1-8}

\multirow{2}{*}{\textbf{ResNet50}}
& \textbf{ThAcc (\%)} & 34.70 & 27.51 & 57.18 & 13.22 & \textbf{92.55} & 90.46 \\
& \textbf{RMSE} & 3.768 & 2.736 & 1.915 & 2.229  & \textbf{0.496} & 0.563 \\
\cline{1-8}

\multirow{2}{*}{\textbf{DenseNet169}}
& \textbf{ThAcc (\%)} & 13.41 & 15.67 & 13.78 & 9.54 & 
 \textbf{94.98} & 74.32 \\
& \textbf{RMSE} & 2.040 & 1.982 & 2.022 & 2.128 & \textbf{0.508} &  0.715 \\
\cline{1-8}

\multirow{2}{*}{\textbf{EfficientNetB0}}
& \textbf{ThAcc (\%)} & 11.66 & 6.32 & 4.91 & 9.78 & 95.65 & \textbf{96.13} \\
& \textbf{RMSE} & 2.261 & 2.659 & 2.731 & 2.283  & \textbf{0.451} & 0.490 \\
\toprule
\end{tabular}
\caption{Evaluation metrics computed between the ground truth and the predicted depth obtained by the proposed method using the testing set of Level 5. Traditional and curriculum learning were tested (columns), with three networks (rows). For curriculum, each level was assessed with the final testing set (Level 5).}
\label{table:appendix_results_synthetic}
\end{table*}


\subsection{Qualitative results}
\label{subappendix_eval_synthetic_quali}

Figure~\ref{fig:qualitative_synthetic_single} shows different depth estimations of the synthetic image (level 5) at the top left panel, while its ground truth depth is at the top right. CL models from Levels 1 to 5 are displayed at the bottom panel from columns 1 to 5, while the $6^{th}$ column shows the depth estimation by TL. Visually, a gradual improvement of the estimated depth maps is observable for the three networks as the CL levels increase. For example, in case of ResNet50 (first row at bottom panel), shadows around haustral folds were wrongly estimated, yet the method gradually improves the estimation when Level 5 is reached. Additionally, depth estimated for the farthest polyp is better with CL than with TL, showing in this case a well-defined structure. In case of DenseNet169 and EfficientNetB0, depth was better approximated as CL progresses, i.e.,  networks estimated a depth very similar in the first levels for the farthest and closest structures. Particularly, EfficientNetB0 was the only network able to estimate shape of the closest polyp with CL and TL. In case of large specular reflections, all networks in both strategies produced proper depth estimations.

\setcounter{figure}{0}
\renewcommand{\thefigure}{B.\arabic{figure}}
\begin{figure*}[ht!]
\centering
\includegraphics[scale=1.05]{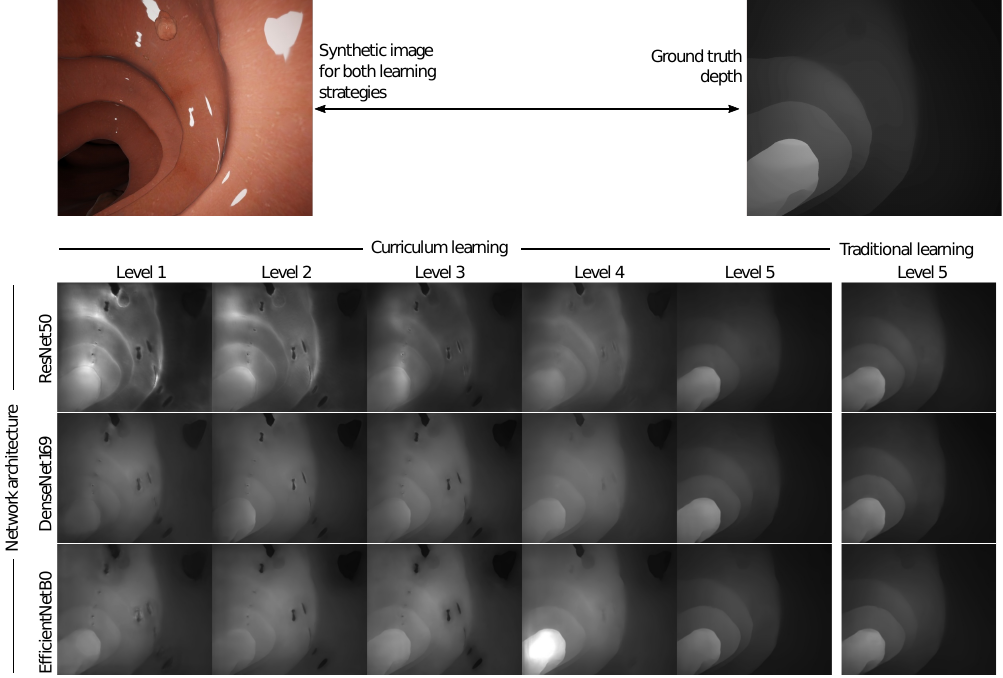}
\caption{For the synthetic frame (top left panel) and its corresponding ground truth (top right panel), Columns 1 to 5 at bottom panel show depth estimations per level of curriculum while the traditional learning is observed in column 6. For each learning strategy, estimations of the three networks are shown: ResNet50 in the first row, DenseNet169 in the second row, and EfficientNetB0 in the third row.}
\label{fig:qualitative_synthetic_single}
\end{figure*}

\bibliographystyle{cas-model2-names.bst}
\bibliography{refs}
\end{document}